\documentclass[11pt]{article}

\usepackage[final]{acl}

\usepackage{times}
\usepackage{latexsym}
\usepackage[T1]{fontenc}
\usepackage[utf8]{inputenc}
\usepackage{microtype}
\usepackage{algorithm}
\usepackage{algorithmic}
\usepackage{graphicx}
\usepackage{subcaption}
\usepackage{booktabs}
\usepackage{multirow}
\usepackage{makecell}
\usepackage{amsmath}
\usepackage{amssymb}
\usepackage{mathtools}
\usepackage{amsthm}
\usepackage{balance}

\usepackage{enumitem}
\setlist[itemize]{noitemsep, topsep=0pt, leftmargin=*}

\usepackage[capitalize,noabbrev]{cleveref}

\usepackage[textsize=tiny]{todonotes}

\theoremstyle{plain}

\theoremstyle{definition}

\theoremstyle{remark}

\title{Trajectory-Triggered Sparse Collaborative Decoding}

\author{
  Amit Ranjan Trivedi\textsuperscript{1}, Sruthikesh Surineni\textsuperscript{2}, Kyong Ho Lee\textsuperscript{2}, \\
  \textbf{Mohit Mittal\textsuperscript{2}, Steven Li\textsuperscript{2}, Rex Hill\textsuperscript{2}} \\
  \textsuperscript{1}University of Illinois Chicago \quad \textsuperscript{2}Meta \\
  \texttt{amitrt@uic.edu}
}

\begin{document}
\maketitle

\begin{abstract}
Dense assistance methods such as speculative decoding repeatedly verify draft tokens with a target model, keeping target computation coupled to generation. We study sparse collaborative decoding, where a draft model generates most tokens and a target model is invoked only for short corrective segments. The key problem is deciding when to call the target. We show that local uncertainty is a weak trigger: entropy, negative log-likelihood, logit margin, top-\(k\) mass, and velocity features give near-chance failure prediction, with AUC \(0.50\)--\(0.53\). In contrast, aggregate trajectory-shape features over the draft confidence path reach AUC \(0.86\) using a lightweight logistic predictor trained only from draft correctness labels. This induces a trigger-timing tradeoff: early calls have more leverage but weaker evidence, while later calls have stronger evidence but less ability to redirect the trajectory. Used online, learned sparse collaborative decoding improves over the best open-loop schedule by \(\mathbf{+5.3}\) to \(\mathbf{+20.0}\) accuracy points across five GSM8K draft-target pairs. On MATH-500, it gains \(7.0\)--\(13.5\) points when early prediction and target competence align, but loses when either is weak. On StrategyQA, it improves over open-loop by \(5.0\)--\(8.3\) points, showing transfer beyond arithmetics. 
\end{abstract}

\section{Introduction}

\begin{figure}[t]
  \centering
  \includegraphics[width=\linewidth]{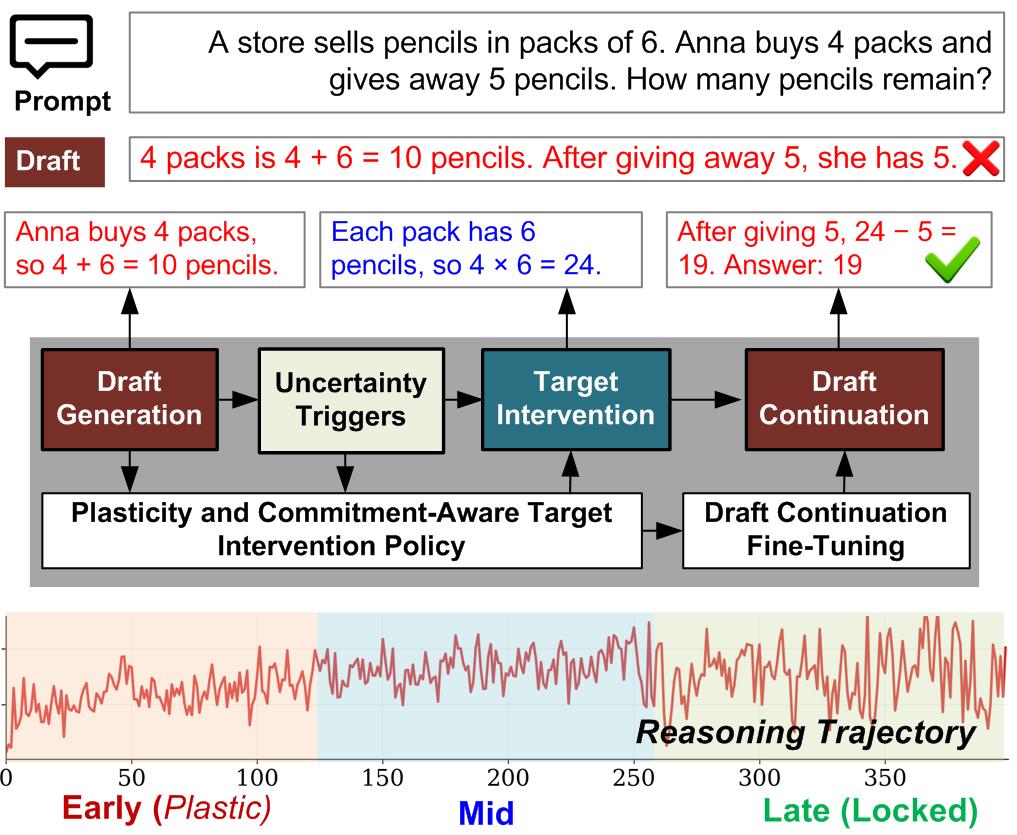}
\caption{\textbf{Trajectory-triggered sparse collaborative decoding.}
A draft model generates most of the reasoning trajectory, while a stronger target model is invoked only at selected points. Triggers use the evolving confidence trajectory rather than fixed schedules or local uncertainty spikes. Early calls have higher leverage but weaker evidence; later calls have less leverage.}\vspace{-15pt}
  \label{fig:introduction}
\end{figure}

Small language models reduce inference cost, latency, and memory use, but remain unreliable on multi-step reasoning tasks. Stronger models are more reliable, but require substantially more storage and computation, motivating small-large collaboration at inference time. Existing methods spend target-model computation coarsely. Speculative decoding repeatedly verifies draft tokens with a target model, preserving target-model behavior but coupling target computation to generation throughout decoding \citep{leviathan2023fast,chen2023accelerating}. Cascades route entire examples to a larger model before generation \citep{chen2023frugalgpt}. These methods either spend target computation continuously or decide only before generation; they do not ask whether the current partial trajectory would benefit from a short target-generated correction. We study \textit{when target computation should be spent inside the trajectory}.

We address this question through \emph{sparse collaborative decoding}. A draft model generates most tokens, while a stronger target model is invoked only for short corrective text segments, after which generation returns to the draft. The intervention is purely textual: the target segment is appended to the visible context, and no logits, hidden states, attention activations, or parameters are modified. This makes the setting compatible with black-box targets and API-based deployment.

The central problem is target-call placement. Calling too early can waste target computation on trajectories the draft would have solved alone. Calling too late leaves little ability to redirect an already committed trajectory. A natural trigger is local uncertainty, such as high entropy, low margin, high negative log-likelihood, or rapidly changing confidence at the current token. We find that this signal is unreliable: across our reasoning settings, per-token uncertainty and velocity features provide near-chance prediction of final failure. A draft can be locally confident on a wrong path, or locally uncertain while still recoverable.

Meanwhile, we find that trajectory shape is a stronger signal. We summarize the draft confidence path using aggregate features of entropy, negative log-likelihood, margin, probability concentration, temporal variation, and position-dependent behavior. A lightweight logistic predictor over these features reaches AUC \(0.86\) for failure prediction. Used online, this predictor estimates failure risk from the partial trajectory and calls the target when risk crosses a threshold. The target then generates a bounded corrective segment, and the draft resumes. This creates a \textit{trigger-timing tradeoff}: early calls have more leverage but weaker evidence, while later calls have stronger evidence but less ability to redirect the trajectory.

Our contributions are threefold. \textit{First}, we show that local uncertainty is a weak intervention trigger, while aggregate confidence-trajectory features provide a much stronger failure signal. \textit{Second}, we convert this signal into an online sparse collaborative decoding policy that decides \emph{where} to spend target computation inside a draft trajectory. \textit{Third}, we evaluate against draft-only, target-only, self-consistency, posthoc rewriting, cascade routing, open-loop SCD, and speculative decoding. Across five GSM8K draft-target pairs, learned triggering improves over the best open-loop schedule by \(5.3\) to \(20.0\) accuracy points. It also transfers to StrategyQA and improves on MATH-500 when early prediction and target competence align. The results show that efficient small-large reasoning depends not only on access to a stronger model, but on placing bounded target calls when they are reliable enough to justify and early enough to matter.

\section{Background and Related Work}
\label{sec:background}

\noindent\textbf{Test-time reasoning and compute allocation.}
Reasoning often improves with additional inference-time computation. Chain-of-thought prompting \citep{wei2022chain,kojima2022large}, self-consistency \citep{wang2022self}, and search-based methods \citep{yao2023tree} spend compute through longer outputs, repeated samples, or branching trajectories. Adaptive-compute methods allocate resources under cost constraints \citep{chen2023frugalgpt,li2025selfbudgeter,yang2025towards}. Comparatively, we ask when a stronger model should be invoked within reasoning trajectory.

\vspace{2pt}
\noindent\textbf{Small-large model collaboration.}
Speculative decoding lets a draft model propose tokens that a target model verifies, improving latency while preserving the target distribution \citep{leviathan2023fast,chen2023accelerating}. Its goal is target-output parity; ours is task accuracy under sparse target access. Cascades route hard examples to larger models before generation \citep{chen2023frugalgpt}, while posthoc rewriting revises only after draft completion. Our method instead decides during generation, using the evolving trajectory.

\vspace{2pt}
\noindent\textbf{Uncertainty and failure prediction.}
Local uncertainty is a natural intervention signal. Entropy, token likelihood, logit margin, and confidence change are widely used in selective prediction, abstention, early exiting, and adaptive inference \citep{geifman2017selective,kadavath2022language,kong2022accelerating,schuster2022confident}. In multi-step reasoning, however, local uncertainty can mislead: a model can be confident on a wrong path or uncertain while still recoverable. We therefore distinguish token-level uncertainty from trajectory-level failure prediction, and show that aggregate confidence paths are more predictive.

\vspace{2pt}
\noindent\textbf{Decoding interventions and trajectory sensitivity.}
Inference-time steering modifies generation through prompts, logits, activations, or continuous prefixes \citep{dathathri2019plug,krause2021gedi,li2021prefix,turner2023steering}. Model editing modifies parameters or internal representations \citep{ilharco2022editing}. Our intervention is textual and sparse: the target produces a short segment, it is appended to the draft context, and the draft resumes. Prior work shows that early prompts and prefixes can strongly affect later generations \citep{zhao2021calibrate,zhou2022least,sun2025well,song2025well}. We study the complementary control problem: early intervention has higher leverage, while later intervention has stronger evidence, motivating learned triggers than schedules.

\section{Problem Setup}
\label{sec:setup}

We consider a draft model \(M_d\) and a stronger target model \(M_t\). Given input \(x\), the draft generates a reasoning trajectory \(y_{1:T}\). The goal is to improve final-answer accuracy while limiting target use.

In \emph{sparse collaborative decoding}, the draft generates most tokens and the target is invoked only at selected intervention points. If a trigger fires after \(y_{1:t}\), the target generates a short segment \(z_{1:J} \sim M_t(\cdot \mid x,y_{1:t})\). This segment is appended to the visible context, and the draft resumes from \((x,y_{1:t},z_{1:J})\). The intervention is purely textual: no logits, hidden states, attention activations, or parameters are modified. This makes the mechanism compatible with black-box targets and distinct from speculative decoding and steering.

Let \(s_t\) denote the observable draft state at step \(t\), including token probabilities, entropy, logit margins, and their history. A trigger policy \(\pi\) maps this state to \(a_t=\pi(s_t)\in\{0,1\}\), where \(a_t=1\) calls the target. The central problem is to design \(\pi\) so that target calls occur when they are likely to improve the final answer, not merely when the draft appears locally uncertain.

We compare seven policies. \emph{Draft-only} uses \(M_d\) throughout, and \emph{target-only} uses \(M_t\) directly. \emph{Open-loop SCD} calls the target at fixed positions or prefix lengths, while \emph{closed-loop SCD} uses the partial draft trajectory to decide when help is needed. \emph{Cascade routing} sends predicted hard examples to the target as whole-example queries. \emph{Posthoc rewriting} asks the target to revise a completed draft answer. \emph{Self-consistency} samples multiple draft trajectories. We report final-answer accuracy, target calls, target-generated tokens, and wall-clock time when available. We also track \emph{rescue rate}, the fraction of draft-wrong examples corrected, and \emph{break rate}, the fraction of correct examples made wrong. 

\section{Trajectory-Level Failure Prediction}
\label{sec:failure_prediction}

A trigger policy must predict whether the draft trajectory is likely to fail. The simplest approach is local uncertainty at the current decoding step. We record entropy \(H_t\), negative log-likelihood \(\mathrm{NLL}_t\), logit margin \(M_t\), top-\(k\) probability mass, and short-window velocity features. These are natural intervention signals, but weak predictors of final correctness. As shown in Table~\ref{tab:failure_prediction}, individual per-token signals on GSM8K with Qwen-1.5B are close to chance: entropy and NLL reach only AUC \(0.531\) and \(0.528\), while velocity features are uninformative. A draft can be locally confident on an incorrect path, or locally uncertain while recoverable.

We therefore predict failure from the shape of the confidence trajectory rather than from a single token. For each base signal, we summarize mean, extrema, percentiles, slope, variance, and position-dependent behavior over the observed draft prefix. The resulting feature vector has about 85 dimensions, with the full list in Appendix~\ref{app:features}. A lightweight logistic predictor over these trajectory-shape features reaches AUC \(0.864\). Strong individual features include top-5 mass statistics, entropy variation, maximum entropy, and NLL percentiles. The largest learned weights involve log-margin slope, absolute margin-velocity slope, log-margin variation, and top-5 mass statistics, indicating that failures are marked by declining separation, increasing volatility, and weaker probability concentration, not by isolated uncertainty spikes.

\begin{table}[t]
\centering
\caption{Failure prediction on GSM8K with Qwen-1.5B. Local token-level signals are weak, while aggregate trajectory-shape features are strongly predictive.}
\label{tab:failure_prediction}
\small
\setlength{\tabcolsep}{5pt}
\begin{tabular}{lcc}
\toprule
Signal or feature set & AUC & Type \\
\midrule
Entropy \(H_t\) & 0.531 & local \\
NLL \(\mathrm{NLL}_t\) & 0.528 & local \\
Margin \(M_t\) & 0.474 & local \\
Entropy velocity & 0.503 & local change \\
NLL velocity & 0.499 & local change \\
Margin velocity & 0.498 & local change \\
\midrule
Best single trajectory feature & 0.803 & trajectory \\
Top trajectory features & 0.747--0.803 & trajectory \\
All trajectory features & \textbf{0.864} & trajectory \\
\bottomrule
\end{tabular}
\end{table}

\begin{figure}[t]
  \centering
  \includegraphics[width=\linewidth]{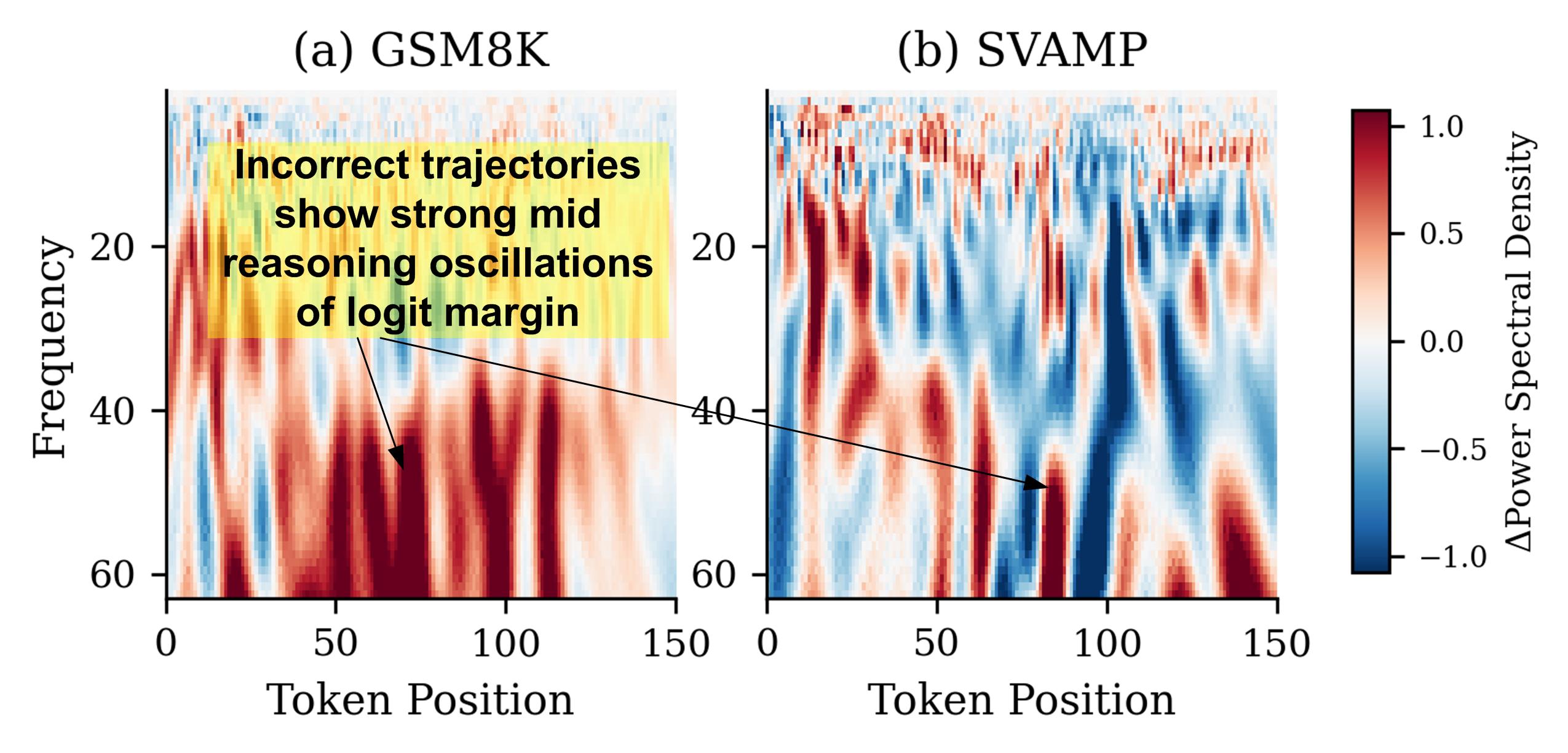}
  \caption{\textbf{Time-frequency view of trajectory instability.}
  Wavelet transforms of logit-margin-derived signals show elevated mid-frequency oscillations for incorrect trajectories, consistent with unresolved competition between reasoning paths before commitment. Correct trajectories show smoother consolidation.}
  \label{fig:wavelet_instability}
\end{figure}

A complementary time-frequency diagnostic supports the same conclusion. 
\Cref{fig:wavelet_instability} shows wavelet transforms of logit-margin-derived signals. Incorrect trajectories show elevated early mid-frequency oscillations, consistent with unresolved competition between reasoning paths, whereas correct trajectories consolidate more smoothly. This supports aggregate trajectory-shape features over one-step uncertainty thresholds.

Prediction quality also depends on when the trigger must fire. On GSM8K, the Qwen-1.5B predictor rises from AUC 0.60 at 16 tokens to 0.82 at 128 tokens and 0.86 on the full trajectory. Table~\ref{tab:early_reliability} shows the broader pattern. GSM8K predictors reach AUC@128 near or above 0.7 across drafts. MATH-500 predictors remain much weaker at the same point, around 0.57--0.59, even though their full-trajectory AUC reaches 0.73--0.78.

\begin{table}[t]
\centering
\caption{Early predictor reliability across draft and dataset pairs. AUC@128 is the operational quantity for online triggering; full AUC is more relevant to posthoc or cascade policies.}
\label{tab:early_reliability}
\small
\setlength{\tabcolsep}{4pt}
\begin{tabular}{llcccc}
\toprule
Draft & Dataset & Fail\% & @16 & @128 & Full \\
\midrule
Qwen 1.5B & GSM8K & 26 & 0.60 & \textbf{0.82} & 0.86 \\
Llama 1B & GSM8K & 52 & 0.61 & \textbf{0.76} & 0.80 \\
Qwen 0.5B & GSM8K & 52 & 0.63 & \textbf{0.69} & 0.78 \\
\midrule
Qwen 0.5B & MATH-500 & 76 & 0.45 & 0.57 & 0.75 \\
Llama 1B & MATH-500 & 86 & 0.56 & 0.59 & 0.73 \\
Qwen 1.5B & MATH-500 & 58 & 0.50 & 0.58 & 0.78 \\
\bottomrule
\end{tabular}
\end{table}

\begin{figure}[t]
  \centering
  \includegraphics[width=\linewidth]{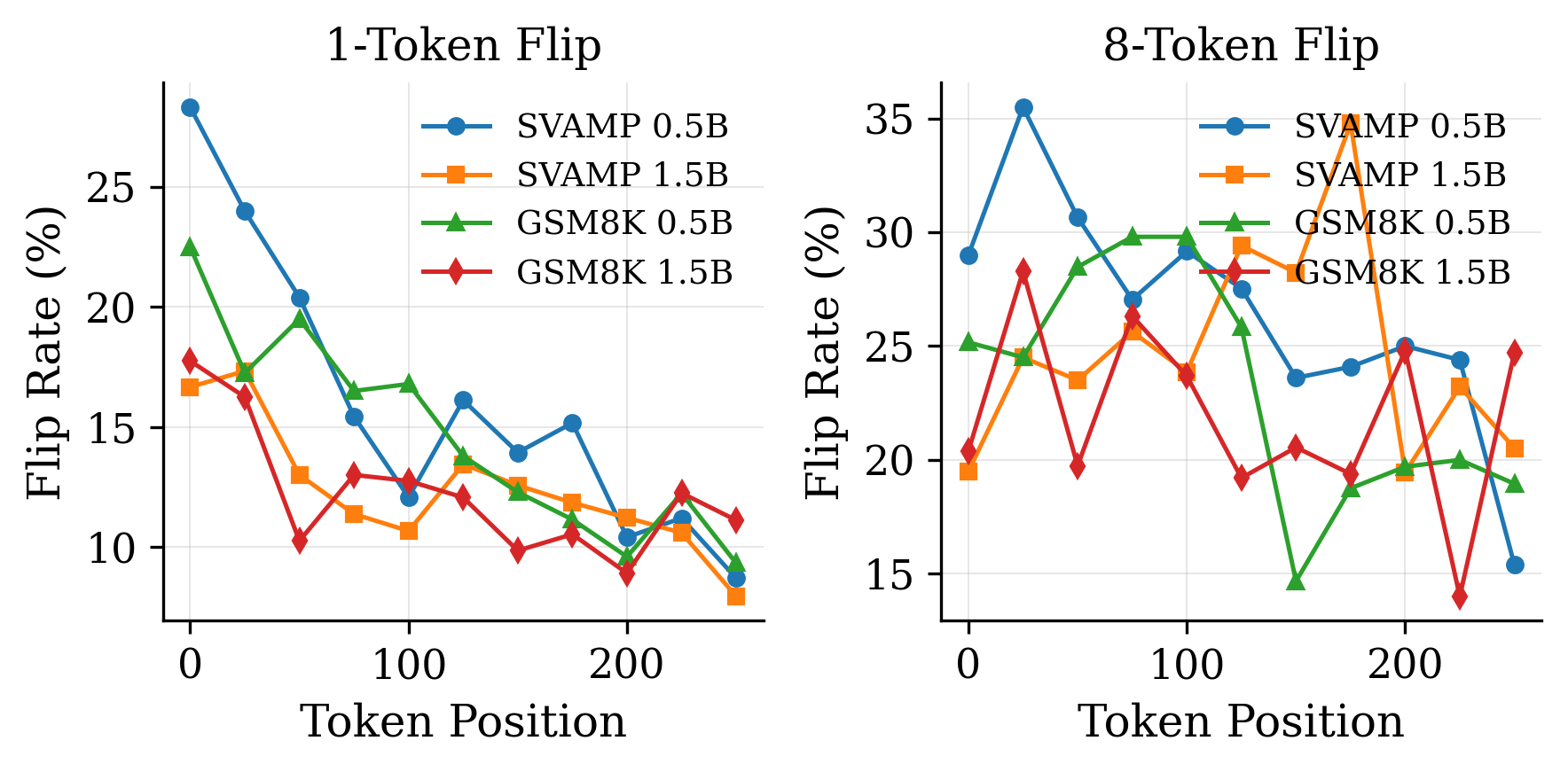}
\caption{\textbf{Intervention leverage decreases with decoding position.}
We perturb the draft prefix at different token positions and measure whether the final decoded sequence changes. Single-token and eight-token perturbations are most influential early and decay with position. Eight-token perturbations raise the absolute flip rate but preserve the same trend. This supports the trigger-timing tradeoff: early calls have greater influence but weaker evidence, while later calls have stronger evidence but less ability to redirect.}
  \label{fig:flip_rates}
\end{figure}

Early reliability conflicts with intervention leverage. We perturb the draft prefix at different token positions and measure the \emph{flip rate}, the fraction of examples whose final decoded sequence changes. Figure~\ref{fig:flip_rates} shows that single-token and eight-token perturbations are most influential early and decay with position. Larger perturbations raise flip rate but preserve this trend. Early tokens have greater downstream influence, while later interventions have stronger evidence but less ability to redirect.

Overall, the predictor supplies a usable but time-dependent control signal where early scores are noisy but actionable, while later scores are more reliable but may arrive after the trajectory has become difficult to redirect. The next section converts this partial-trajectory predictor into an online trigger that decides when to spend a bounded number of target calls and target-generated tokens.

\section{Learning Sparse Intervention Triggers}
\label{sec:method}

We convert the trajectory predictor into an online target-call policy. A small calibration set is decoded with the draft model to collect step-level statistics and final-answer correctness labels. We train a logistic predictor \(g_\psi\) on the trajectory-shape features from \Cref{sec:failure_prediction}. At test time, the same feature extractor is applied to the observed trajectory.

At step \(t\), let \(\phi_t=\phi(x,y_{1:t})\) denote features computed from the draft confidence path. The trigger score is \(r_t=g_\psi(\phi_t)\), where larger values indicate higher predicted failure risk. The controller calls the target when \(r_t \ge \theta\), subject to warmup \(W\), cooldown \(C\), and target-call budget \(B\). The threshold \(\theta\) may be fixed or position-dependent to account for the tradeoff between early leverage and later evidence. To prevent cascading trigger loops, where a low-confidence target resumption immediately re-fires the policy, the controller enforces token cooldown \(C\) and target-call budget \(B\) per trajectory. Thus, once control returns to the draft, the trajectory must progress through a mandatory evaluation window before the predictor fires again.

When the trigger fires, the target generates a bounded text segment \(z_{1:J} \sim M_t(\cdot \mid x,y_{1:t})\), where \(J\) is the per-call segment length. The segment is appended to the visible context, and the draft resumes from \((x,y_{1:t},z_{1:J})\). The intervention remains purely textual: no logits, hidden states, attention, or model parameters are modified.

The predictor estimates draft failure only; it does not use target outputs, rewards, or intervention outcomes. This separates trigger quality from target behavior: the trigger must find states where help is useful, and the target must improve the fired context. The main deployment knobs are \(J\), \(B\), and \(\theta\). Segment length \(J\) trades cost for corrective reach, budget \(B\) limits repeated target use, and threshold \(\theta\) controls selectivity. Effective policies are therefore not simply aggressive; they wait for enough trajectory-level evidence while leaving enough remaining generation to affect the answer.

\section{Experiments}
\label{sec:experiments}

\noindent\textbf{Setup.}
We evaluate on GSM8K \citep{cobbe2021training}, MATH-500 \citep{hendrycks2021measuring}, and AIME-2024 using exact-match answer accuracy. GSM8K is the main controlled setting, MATH-500 tests harder symbolic reasoning, and AIME-2024 is a stress case where target models are near floor. Drafts include Qwen2.5-0.5B and Qwen2.5-1.5B \citep{qwen2025technical}, and Llama-3.2-1B \citep{grattafiori2024llama}; targets are gpt-4o-mini and gpt-4o. Draft decoding is greedy.\footnote{Each result cell uses $n=200$ examples (except StrategyQA at $n=300$ and the Q9 generalization triples at $n=200$) and a single fixed seed; the binomial standard error at $n=200$ is at most $\pm 3.5$pp at $50\%$ accuracy and $\pm 1.6$pp at $90\%$, so the $+5$pp closed-loop-vs-open-loop gaps we report are well outside the single-seed noise band. Reported triples reuse the same example list across methods for paired comparison.} For each draft-dataset pair, we collect draft-only logs, train the trajectory-shape predictor from \Cref{sec:failure_prediction}, and deploy the learned trigger without using target outputs.

We compare seven policies: draft-only, target-only, self-consistency, posthoc rewrite, cascade routing, open-loop SCD, and learned-trigger closed-loop SCD. Open-loop SCD uses the same text-injection primitive as our method but fixes the insertion schedule. Cascade routes predicted hard examples to the target as whole-example queries. Posthoc rewrite applies the predictor after draft completion and asks the target to revise. The appendix reports full \(K \times P\) open-loop heatmaps.

\begin{table}[t]
\centering
\caption{Open-loop SCD on GSM8K. Per-\(K\) marginals average over schedule positions \(P\). Best cell reports the best position for each prefix length.}
\label{tab:openloop_marginals}
\small
\setlength{\tabcolsep}{3.5pt}
\begin{tabular}{l r r r r}
\toprule
Draft & \(K\) & Mean acc & Tgt tok & Best cell \\
\midrule
Qwen 1.5B & 8   & 75.8 & 8.3   & 77.2 \\
Qwen 1.5B & 32  & 78.6 & 33.2  & 81.5 \\
Qwen 1.5B & 64  & 80.5 & 65.9  & 83.5 \\
Qwen 1.5B & 128 & 84.9 & 126.2 & \textbf{86.5} \\
\midrule
Llama 1B & 8   & 48.8 & 8.4   & 50.0 \\
Llama 1B & 32  & 55.4 & 30.9  & 58.8 \\
Llama 1B & 64  & 60.6 & 59.6  & 63.2 \\
Llama 1B & 128 & 70.8 & 110.8 & \textbf{75.0} \\
\bottomrule
\end{tabular}
\end{table}

\vspace{3pt}
\noindent\textbf{Open-loop SCD is a strong baseline.}
Open-loop SCD is the closest baseline because it uses the same target text insertion as learned-trigger SCD; only call placement differs. \Cref{tab:openloop_marginals} shows that fixed target insertion already gives large gains on GSM8K: Qwen-1.5B reaches \(86.5\%\), and Llama-1B reaches \(75.0\%\). However, rigid schedules can also harm weaker drafts. Blindly injecting \(J=32\) target tokens drops Llama-1B mean accuracy to \(30.9\%\), suggesting that fixed insertions can disrupt the draft's internal plan. Learned triggering is therefore evaluated against a strong target-prefix baseline, not only against draft-only decoding.

Open-loop spends target tokens on every example, including trajectories the draft would have solved alone, and cannot adapt when different examples drift at different positions. Learned triggering keeps the same target-text primitive but decides which trajectories and positions receive target help.

\begin{table}[t]
\centering
\caption{Closed-loop J-sweeps on GSM8K. Each row reports the best accuracy at that segment length over tested thresholds.}
\label{tab:closedloop_jsweep_compact}
\small
\setlength{\tabcolsep}{3.5pt}
\begin{tabular}{l r r r r r}
\toprule
Draft & \(J\) & Best acc & Rescue & Break & Tgt tok \\
\midrule
Qwen 1.5B & 8   & 80.8 & 10.8 & 4.5 & 33.9 \\
Qwen 1.5B & 32  & 87.5 & 15.2 & 2.2 & 97.6 \\
Qwen 1.5B & 64  & 91.0 & 17.5 & 1.0 & 122.3 \\
Qwen 1.5B & 96  & \textbf{91.8} & 18.2 & 1.0 & 141.3 \\
\midrule
Llama 1B & 8   & 50.5 & 11.8 & 9.0 & 49.4 \\
Llama 1B & 32  & 77.8 & 33.2 & 3.2 & 151.7 \\
Llama 1B & 64  & 85.0 & 40.0 & 2.8 & 182.9 \\
Llama 1B & 96  & \textbf{88.8} & 42.5 & 1.5 & 195.1 \\
\bottomrule
\end{tabular}
\end{table}

\begin{figure}[t]
  \centering
  \includegraphics[width=0.48\linewidth]{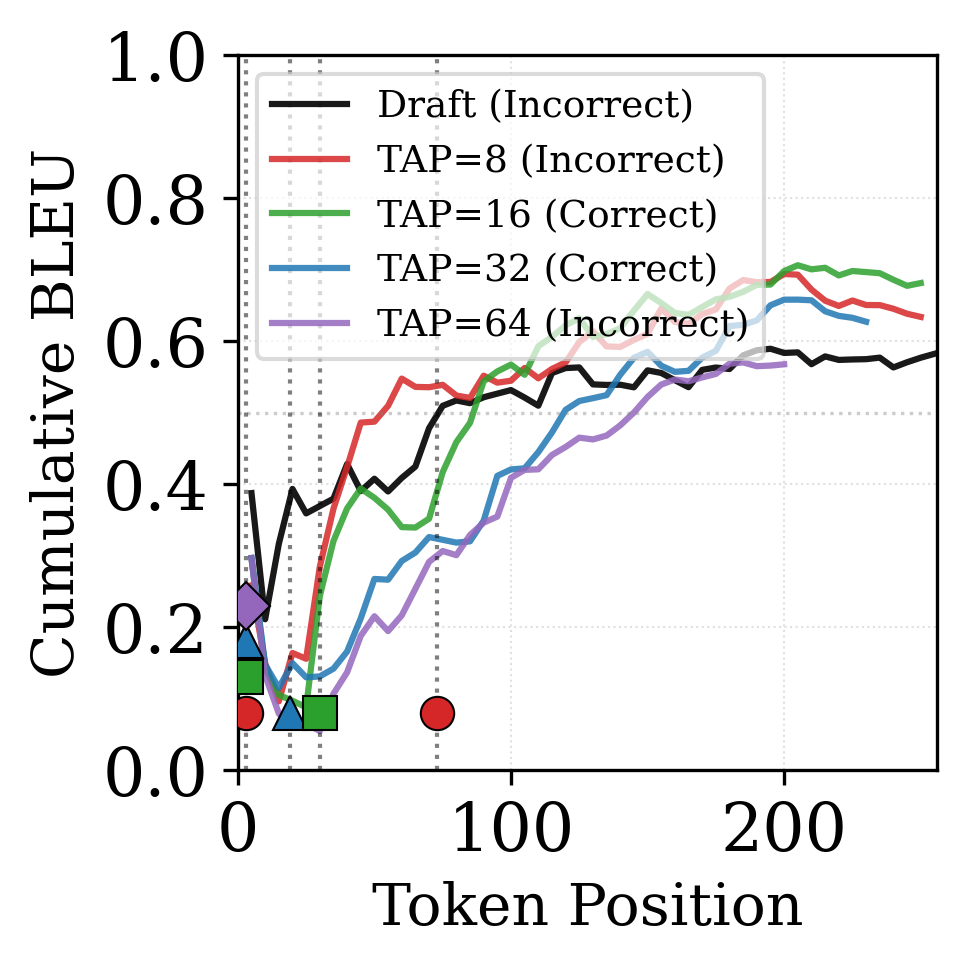}
  \includegraphics[width=0.48\linewidth]{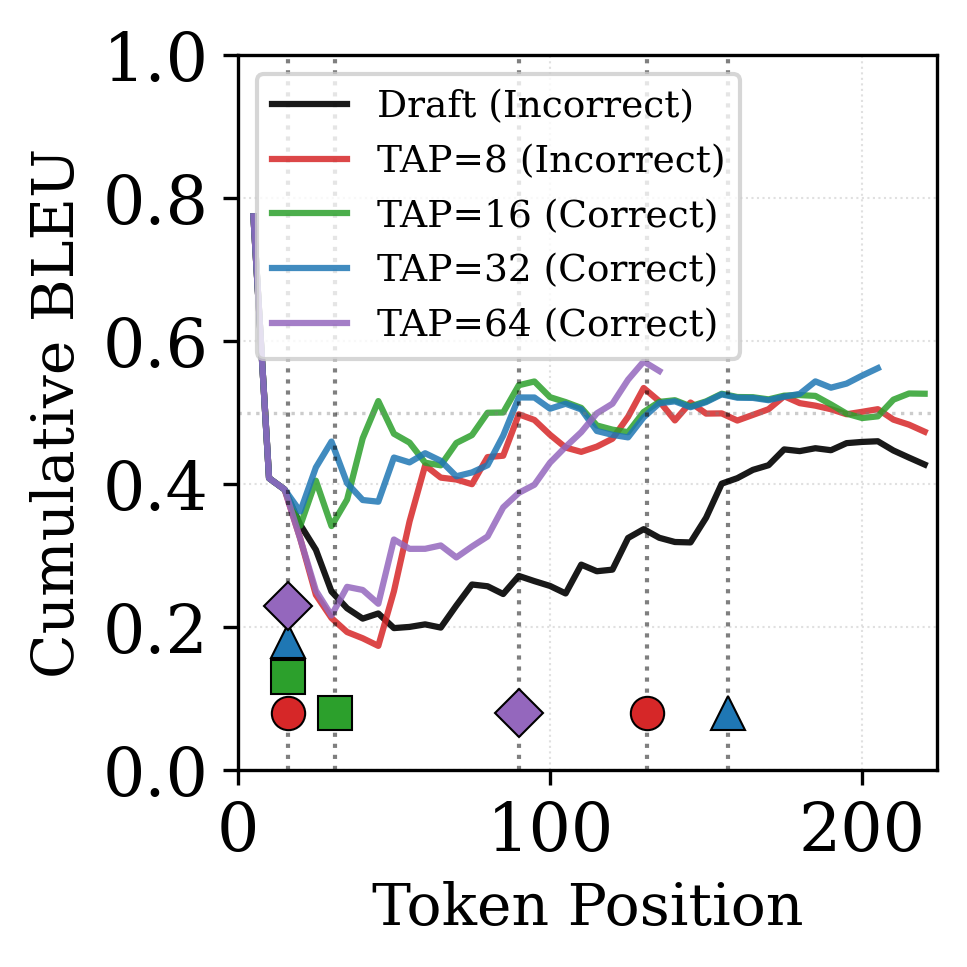}
\caption{\textbf{Closed-loop intervention recovers trajectory alignment.}
We plot cumulative BLEU \citep{papineni2002bleu} between the draft trajectory and a target-guided reference. Draft-only decoding drifts and does not recover. Triggered target insertions occur after alignment degradation and induce rapid recovery, showing that sparse calls can redirect the trajectory.}
  \label{fig:anytime_bleu}
\end{figure}

\vspace{3pt}
\noindent\textbf{Learned triggers improve the accuracy-cost frontier.}
\Cref{tab:closedloop_jsweep_compact} summarizes closed-loop \(J\)-sweeps on GSM8K. Segment length \(J\) controls intervention reach: very short target segments are cheap but often incomplete, while longer segments improve accuracy until target use saturates. For Qwen-1.5B, closed-loop beats the interpolated open-loop frontier in 16 of 21 tested cells, ties in 3, and loses in 2. For Llama-1B, it wins in 13 of 21 cells. Losses occur mainly at small \(J\), where fired target segments are too short to repair the trajectory. The Llama gain is larger because its draft baseline is weaker. We also test whether target calls redirect the trajectory rather than merely adding helpful tokens near the end. \Cref{fig:anytime_bleu} plots representative cumulative alignment traces. Draft-only decoding drifts away from the target-guided trajectory and does not recover. Triggered insertions occur after alignment degradation and produce rapid recovery, supporting the mechanism behind the \(J\)-sweep.

\vspace{3pt}
\noindent\textbf{Thresholds control selectivity and cost.}
The trigger threshold controls the accuracy-cost point. Lower thresholds fire more often and use more target tokens; higher thresholds save target use but miss recoverable failures. We therefore also test a position-dependent schedule, where \(\theta(t)\) linearly interpolates from an early threshold \(\theta_e\) to a late threshold \(\theta_l\). This allows the controller to use different selectivity before and after more trajectory evidence has accumulated. \Cref{tab:posdep_compact} shows representative sweeps at \(J=96\).

\begin{table}[t]
\centering
\caption{Position-dependent threshold sweeps on GSM8K with \(J=96\). \(\theta_e\) and \(\theta_l\) are early- and late-region thresholds. Lower thresholds fire more often, increasing rescue and target use.}
\label{tab:posdep_compact}
\small
\setlength{\tabcolsep}{4pt}
\begin{tabular}{l c c r r r}
\toprule
Draft & \(\theta_e\) & \(\theta_l\) & Acc & Rescue & Tgt tok \\
\midrule
Qwen 1.5B & 0.5 & 0.3 & \textbf{92.2} & 19.0 & 151.6 \\
Qwen 1.5B & 0.5 & 0.5 & 90.8 & 18.0 & 135.1 \\
Qwen 1.5B & 0.7 & 0.7 & 88.0 & 14.8 & 90.7 \\
\midrule
Llama 1B & 0.3 & 0.5 & \textbf{90.0} & 44.0 & 232.0 \\
Llama 1B & 0.5 & 0.5 & 86.0 & 41.2 & 195.7 \\
Llama 1B & 0.7 & 0.7 & 77.2 & 32.0 & 136.9 \\
\bottomrule
\end{tabular}
\end{table}

The effect is draft-dependent. On Qwen-1.5B, position dependence is directionally useful but small, with gains over the constant \(\theta=0.5\) setting within standard error in the full sweep. On Llama-1B, threshold choice matters much more where aggressive early firing yields large gains because the weaker draft has many more recoverable failures. Thus \(\theta_e,\theta_l\) should be treated as deployment knobs, not universal constants.

\vspace{3pt}
\noindent\textbf{Generalization across tasks.}
\Cref{tab:generalization_main} evaluates math reasoning and target-floor regimes. Learned-trigger SCD wins on all five GSM8K triples, improving over the best open-loop schedule by $+5.3$ points (Qwen-1.5B + gpt-4o-mini) to $+20.0$ points (Qwen-0.5B + gpt-4o-mini). The gain scales inversely with draft capacity: smaller drafts have more recoverable failures, so a well-timed target call closes a larger fraction of the draft-to-target gap. This is a deployment-relevant property: the method's value grows precisely in the small-model regime that motivates sparse collaboration. On MATH-500, it wins on three of five triples, with \(+7.0\) to \(+13.5\) point gains. The two MATH losses occur for Qwen drafts paired with gpt-4o-mini. As shown in the appendix predictor-reliability table, operational \(\textsc{auc}@128\) on MATH-500 remains near chance (\(0.57\)--\(0.59\)), well below GSM8K (\(0.76\)--\(0.82\)). Because symbolic math evidence accumulates more slowly, the trigger fires from noisier trajectory data and hands compromised contexts to a target that cannot reliably repair them. On AIME-2024, target models are near floor, leaving little capability to allocate.

\begin{table}[t]
\centering
\caption{Generalization across math and stress-test triples. Parentheses report average target-generated tokens per example.}
\label{tab:generalization_main}
\small
\setlength{\tabcolsep}{3pt}
\begin{tabular}{llccc}
\toprule
Draft & Target & D-only & Best OL & Best CL \\
\midrule
\multicolumn{5}{l}{\emph{GSM8K}} \\
Qwen 1.5B & 4o-mini & 74.5 & 86.5 (104) & \textbf{91.8} (141) \\
Llama 1B  & 4o-mini & 47.8 & 75.0  (91) & \textbf{88.8} (195) \\
Qwen 0.5B & 4o-mini & 48.0 & 69.5 (118) & \textbf{89.5} (224) \\
Qwen 0.5B & 4o      & 48.0 & 72.5 (116) & \textbf{91.0} (196) \\
Llama 1B  & 4o      & 45.0 & 68.5  (66) & \textbf{86.5} (192) \\
\midrule
\multicolumn{5}{l}{\emph{MATH-500}} \\
Llama 1B  & 4o-mini & 14.0 & 19.5 (118) & \textbf{30.5} (620) \\
Llama 1B  & 4o      & 14.0 & 21.5 (115) & \textbf{35.0} (562) \\
Qwen 0.5B & 4o      & 24.5 & 34.5 (129) & \textbf{41.5} (536) \\
Qwen 0.5B & 4o-mini & 24.5 & \textbf{37.5} (130) & 32.0 (612) \\
Qwen 1.5B & 4o-mini & 42.5 & \textbf{46.5} (129) & 42.0 (525) \\
\bottomrule
\end{tabular}
\end{table}

The MATH rows isolate the target-competence limit. Qwen-0.5B MATH uses the same draft and predictor in both target settings. Replacing gpt-4o-mini with gpt-4o flips closed-loop from a \(-5.5\) point loss to a \(+7.0\) point gain. Thus failure is not explained by the trigger alone: early firing must be paired with a target that can repair the fired context.

\vspace{3pt}
\noindent\textbf{Generalization to commonsense reasoning.}
To test whether gains are specific to arithmetic, we evaluate on StrategyQA \citep{geva2021did}, a multi-hop commonsense yes/no benchmark with shorter trajectories and no numeric answer extraction. \Cref{tab:strategyqa} shows that closed-loop SCD wins on all three gpt-4o-mini draft-target triples, improving over the best open-loop cell by \(+5.0\) to \(+8.3\) points and over draft-only by \(+8.3\) to \(+20.4\) points. The strongest transfer case is Llama-1B, where closed-loop improves from a \(48.3\%\) draft baseline to \(68.7\%\), adding \(+8.0\) points over open-loop. Because StrategyQA trajectories are shorter than GSM8K, we scale the warmup window \(W\) and baseline threshold sequence proportionally to the task's maximum sequence length, without manual feature re-engineering.

\begin{table}[t]
\centering
\caption{Generalization to commonsense multi-hop QA on StrategyQA (\(n=300\)). All rows use gpt-4o-mini.}
\label{tab:strategyqa}
\small
\setlength{\tabcolsep}{4pt}
\begin{tabular}{lrrrr}
\toprule
Draft & D-only & Best OL & Best CL & \(\Delta\) \\
\midrule
Qwen 0.5B & 50.0 & 50.0 & \textbf{58.3} & \textbf{+8.3} \\
Llama 1B  & 48.3 & 60.7 & \textbf{68.7} & \textbf{+8.0} \\
Qwen 1.5B & 52.3 & 62.3 & \textbf{67.3} & \textbf{+5.0} \\
\bottomrule\vspace{-20pt}
\end{tabular}
\end{table}

\vspace{2pt}
\noindent\textbf{Posthoc, cascade, and standard baselines.}
Posthoc rewrite tests whether target help can be delayed until draft completion. It is dominated by in-trajectory intervention: the best posthoc setting reaches \(76.2\%\) for Qwen-1.5B and \(59.8\%\) for Llama-1B, versus \(91.8\%\) and \(88.8\%\) for closed-loop SCD. Short rewrite budgets can also break correct drafts through incomplete revisions.

Cascade is stronger because it routes predicted hard examples directly to the target. On Llama-1B with gpt-4o-mini on GSM8K, cascade reaches \(87.8\%\), close to closed-loop at \(88.8\%\), showing that the predictor explains much of the gain. The remaining gain comes from intervening inside the draft trajectory rather than discarding draft computation. \Cref{tab:baselines_main} compares all seven methods on this triple. Self-consistency drops accuracy from \(47.8\%\) to \(46.5\%\). Open-loop confirms that target text helps, but learned triggering adds another \(+13.8\) points. Closed-loop recovers about \(89\%\) of the draft-to-target gap, reaching \(88.8\%\) against a \(93.8\%\) target-only reference. Its wall-clock time is \(7.8\)s versus \(6.8\)s for target-only because API deployment requires multiple network round-trips and generation resets. With local weights or unified serving, this handshake overhead should decrease.

The \(+1.0\) point over cascade is small but structurally meaningful. Cascade commits the whole example to the target once flagged, discarding draft computation and paying the full target sequence length. Closed-loop SCD keeps the draft as the default generator and pays only for bounded fired segments: cascade uses \(1.00\) target call of mean length \(184.8\) tokens, whereas closed-loop uses \(2.27\) shorter calls averaging \(86\) tokens each. When the failing region is only part of the trajectory, as on most GSM8K cases, in-trajectory intervention is preferable to whole-example escalation. Cascade and closed-loop share the trigger; they differ in \emph{what} they hand to the target.

\begin{table}[t]
\centering
\caption{Seven-method comparison on Llama-1B with gpt-4o-mini on GSM8K. Closed-loop SCD gives the best accuracy below target-only.}
\label{tab:baselines_main}
\small
\setlength{\tabcolsep}{4pt}
\begin{tabular}{l r r r r}
\toprule
Method & Acc & Tgt tok & Calls & Wall s \\
\midrule
Draft-only & 47.8 & 0.0 & 0.00 & 9.0 \\
Target-only & 93.8 & 274.5 & 1.00 & 6.8 \\
Self-consistency \(k=5\) & 46.5 & 0.0 & 0.00 & 49.4 \\
Posthoc rewrite & 59.8 & 147.6 & 0.61 & 10.0 \\
Open-loop SCD & 75.0 & 90.6 & 0.90 & 6.3 \\
Cascade & 87.8 & 184.8 & 0.61 & 10.5 \\
\textbf{Closed-loop SCD} & \textbf{88.8} & 195.1 & 2.27 & 7.8 \\
\bottomrule
\end{tabular}
\end{table}

\vspace{3pt}
\noindent\textbf{Failure modes and deployment rule.}
Closed-loop failures expose two bottlenecks. The first is predictor-side: when early AUC is weak, the trigger fires with too little evidence or after the draft has already committed. In this regime, the same signal is better suited to cascade or posthoc routing. The second is target-side: even a well-timed trigger can hurt if the target continuation is uncertain, too long, or inconsistent with the draft's current plan. Such segments overwrite rather than correct.

\begin{table}[t]
\centering
\caption{\textbf{Predictor-side failure mode.} Triples where the early trigger is unreliable lose under closed-loop SCD, while triples with $\textsc{auc}@128 \ge 0.70$ win. Pairs identical drafts on different targets isolate the predictor effect (rows~3-4).}
\label{tab:predictor_failure_mode}
\small
\setlength{\tabcolsep}{1.5pt}
\begin{tabular}{llccr}
\toprule
Draft & Dataset & \textsc{auc}@128 & Full \textsc{auc} & $\Delta_{\text{CL-OL}}$ \\
\midrule
Qwen 1.5B & GSM8K     & \textbf{0.82} & 0.86 & $+5.3$ \\
Llama 1B  & GSM8K     & \textbf{0.76} & 0.80 & $+13.8$ \\
Qwen 0.5B & GSM8K     & 0.69          & 0.78 & $+20.0$ \\
\midrule
Qwen 0.5B & MATH-500  & 0.57          & 0.75 & $-5.5$ {\scriptsize(4o-mini)} \\
Qwen 0.5B & MATH-500  & 0.57          & 0.75 & $+7.0$ {\scriptsize(4o)} \\
Qwen 1.5B & MATH-500  & 0.58          & 0.78 & $-4.5$ \\
Llama 1B  & MATH-500  & 0.59          & 0.73 & $+11.0$ \\
\bottomrule
\end{tabular}
\end{table}

\Cref{tab:predictor_failure_mode} shows the predictor-side picture: every triple with $\textsc{auc}@128 \ge 0.70$ wins, every triple below it relies on the target gate to recover. The Qwen-0.5B MATH-500 row appears twice because the same predictor is paired with two targets: holding the predictor fixed and only swapping gpt-4o-mini $\to$ gpt-4o flips a $-5.5$pp loss into a $+7.0$pp gain, confirming both gates are needed.

\begin{table}[t]
\centering
\caption{Target behavior at intervention time. Degrading interventions are less confident and more uncertain than recovery interventions.}
\label{tab:target_uncertainty_main}
\small
\setlength{\tabcolsep}{4pt}
\begin{tabular}{lcc}
\toprule
Metric & Recovery & Degradation \\
\midrule
Mean NLL \(\downarrow\) & 0.162 & 0.200 \\
Mean entropy \(\downarrow\) & 0.336 & 0.413 \\
Mean max prob \(\uparrow\) & 0.877 & 0.854 \\
Mean margin \(\uparrow\) & 0.795 & 0.765 \\
Hedging terms & 4.1 & 38.2 \\
Incomplete computation & 0.8 & 22.0 \\
\bottomrule
\end{tabular}
\end{table}

\Cref{tab:target_uncertainty_main} shows the target-side effect directly: degrading interventions have higher NLL and entropy, lower maximum probability and margin, and more uncertain text. Thus closed-loop failure is not only trigger error; the target must also produce bounded, reliable continuations. \Cref{tab:target_ladder_main} further shows that target-only accuracy is insufficient. Holding the draft, predictor, and closed-loop parameters fixed, gpt-4o gives the best closed-loop accuracy even though gpt-4.1 has higher target-only accuracy. The difference is intervention behavior: gpt-4o produces shorter continuations, while the gpt-4.1 family produces longer segments that overwrite more draft context and degrade correct drafts.

\begin{table}[t]
\centering
\caption{Target capability is not enough. Same draft, predictor, and closed-loop parameters; only the target changes. Tok is mean target tokens per intervention.}
\label{tab:target_ladder_main}
\small
\setlength{\tabcolsep}{3pt}
\begin{tabular}{llrrrr}
\toprule
Data & Target & T-only & CL & Deg. & Tok \\
\midrule
GSM8K & 4o-mini & 94.5 & 87.0 & 1.5 & 142.7 \\
GSM8K & 4.1-mini & \textbf{97.5} & 84.0 & 4.5 & 240.9 \\
GSM8K & 4o & 95.0 & \textbf{91.0} & 2.0 & \textbf{122.9} \\
GSM8K & 4.1 & 96.5 & 79.5 & 7.0 & 346.2 \\
\midrule
MATH & 4o-mini & 69.5 & 49.0 & 5.0 & 405.0 \\
MATH & 4.1-mini & \textbf{72.0} & 37.5 & 12.0 & 493.4 \\
MATH & 4o & 71.0 & \textbf{53.5} & \textbf{4.5} & \textbf{360.5} \\
MATH & 4.1 & 71.5 & 40.5 & 10.5 & 482.1 \\
\bottomrule
\end{tabular}
\end{table}

These results suggest a simple two-knob deployment rule. Use closed-loop SCD when both hold:
\begin{equation}
\label{eq:deploy_rule}
\underbrace{\textsc{auc}@128 \;\ge\; 0.70}_{\text{predictor gate}} \quad\text{and}\quad \underbrace{\frac{\bar{\tau}_{\mathrm{intv}}}{\bar{\tau}_{\mathrm{target\text{-}only}}} \;\le\; 1.2}_{\text{target gate}} ,
\end{equation}
where $\bar{\tau}_{\mathrm{intv}}$ is mean target tokens per intervention and $\bar{\tau}_{\mathrm{target\text{-}only}}$ is mean standalone target completion length. If the predictor gate fails, the same risk score should drive \emph{cascade} or \emph{posthoc} routing instead. If the target gate fails, use a terser target, reduce $J$, or fall back to open-loop SCD. Across the $14{+}3$ deployment triples in \Cref{tab:generalization_main,tab:strategyqa}, \Cref{eq:deploy_rule} selects gpt-4o for every winning cell and correctly rules out the two MATH losses.

\begin{table}[t]
\centering
\caption{SD vs. SCD on SD-feasible Qwen pairs. Cells report accuracy / target tokens. Target is Qwen-14B.}
\label{tab:sd_compare_main}
\scriptsize
\setlength{\tabcolsep}{2.5pt}
\begin{tabular}{lcccc}
\toprule
Method & \multicolumn{2}{c}{GSM8K} & \multicolumn{2}{c}{MATH} \\
\cmidrule(lr){2-3}\cmidrule(lr){4-5}
 & 1.5B & 0.5B & 1.5B & 0.5B \\
\midrule
Draft & 73.5 / 0 & 48.0 / 0 & 50.0 / 0 & 29.5 / 0 \\
OL-SCD & 83.5 / 87 & 69.5 / 109 & 57.5 / 95 & 38.0 / 125 \\
CL-SCD$_{\rm low}$ & 87.5 / 120 & 88.5 / 409 & 55.5 / 198 & 27.0 / 336 \\
CL-SCD$_{\rm best}$ & 92.0 / 330 & 91.0 / 501 & -- & -- \\
SD & 93--95 / 244 & 94--95 / 244 & 70.5--74 / 556 & 70.5--71.5 / 558 \\
Target & 94.0 / 1280 & 93.5 / 512 & 48.0 / 507 & 48.0 / 507 \\
\bottomrule
\end{tabular}
\vspace{-6pt}
\end{table}
\vspace{2pt}

\vspace{2pt}
\noindent\textbf{Predictor and target gates are both necessary.}
The failures are not explained by trigger quality alone. In \Cref{tab:predictor_failure_mode}, the same Qwen-0.5B MATH predictor has low \(\textsc{auc}@128\) in both target settings, yet swapping gpt-4o-mini for gpt-4o flips closed-loop SCD from a \(-5.5\) point loss to a \(+7.0\) point gain. Conversely, \Cref{tab:target_uncertainty_main,tab:target_ladder_main} shows that higher target-only accuracy does not guarantee better intervention: models with longer or more uncertain continuations degrade more correct drafts. Thus sparse collaboration requires a reliable early trigger and a bounded target response, not simply a stronger target model.

\vspace{2pt}
\noindent\textbf{Relation to speculative decoding.}
Speculative decoding (SD) is the closest dense draft-target baseline, but it optimizes a different regime. SD preserves target-output parity through per-token rejection sampling, which requires shared tokenization and access to the target model's full per-token distribution \citep{leviathan2023fast,chen2023accelerating}. These requirements exclude our main API-target setting, where gpt-4o-mini and gpt-4o return text but not full target logits. We therefore compare to SD only where it is well-defined: open-weight Qwen draft-target pairs with shared tokenization and full logit access. \Cref{tab:sd_compare_main} shows the resulting separation. In this SD-feasible regime, SD is the strongest parity-preserving baseline and approaches target-level accuracy through dense verification. SCD instead provides sparse text-correction operating points that do not use target logits. It can run below the full-verification budget, while SD has no partial analogue because it must verify the generated sequence. More importantly, the API-target triples in our main experiments fall outside SD's assumptions, making SCD the applicable mechanism for selective black-box target assistance.

\section{Conclusion}
\label{sec:conclusion}

We studied sparse collaborative decoding, where a draft model generates most tokens and a stronger target model provides short corrective segments. The central problem is target-call placement. Local uncertainty is a weak failure predictor, while aggregate trajectory-shape features provide a stronger online signal. Learned triggers improve over fixed schedules on GSM8K and transfer to StrategyQA when early prediction and target competence align. The results show that sparse collaboration requires both a reliable early trigger and bounded target continuations. SCD offers a trajectory-aware alternative to dense verification, allocating target computation only when it can change the reasoning.

\newpage

\section{Limitations}
\label{app:limitations}

Sparse collaborative decoding has several limitations. \textit{First}, the trigger is task-, draft-, and distribution-dependent. A predictor calibrated on one reasoning distribution may not transfer unchanged to harder tasks, longer contexts, or domains with different error structure. This is visible on MATH-500, where full-trajectory prediction remains useful but early prediction is weaker, making online intervention less reliable. In such cases, the same risk signal may be better suited to cascade or posthoc routing. \textit{Second}, target behavior matters as much as target capability. High target-only accuracy does not guarantee useful intervention: long, uncertain, hedged, or inconsistent continuations can overwrite rather than correct the draft trajectory. Deployment therefore requires calibration of threshold, segment length, cooldown, call budget, and target choice. \textit{Third}, SCD is not a replacement for speculative decoding. When shared tokenization and full target logits are available, and the objective is target-output parity or latency reduction, speculative decoding remains the appropriate dense verification method. SCD instead targets selective black-box or API-based assistance, where target logits are unavailable and task accuracy under limited target access is the main goal. \textit{Fourth}, the evaluation is limited to exact-match reasoning benchmarks. Answer extraction and normalization can affect symbolic datasets, and the results may not directly transfer to long-form generation, dialogue, tool use, multilingual tasks, or domains where errors are not captured by final-answer accuracy. \textit{Finally}, the controller is intentionally lightweight. We use aggregate trajectory-shape features and a logistic predictor to isolate the value of trajectory-level failure prediction. Richer policies could jointly optimize trigger timing, segment length, target selection, and stopping behavior, but would introduce more training cost, hyperparameter sensitivity, and risk of overfitting.

\paragraph{Potential risks.}
Sparse collaborative decoding can increase user trust in outputs that remain incorrect, especially when target interventions appear to repair a reasoning trajectory but do not fix the final answer. It may also increase reliance on black-box target models whose behavior is difficult to audit. In high-stakes settings, SCD should therefore be used with external validation, calibrated abstention, and task-specific safety checks rather than as an automatic correctness guarantee.

\bibliography{refs}

\begin{thebibliography}{28}
\providecommand{\natexlab}[1]{#1}

\bibitem[{Chen et~al.(2023{\natexlab{a}})Chen, Borgeaud, Irving, Lespiau, Sifre, and Jumper}]{chen2023accelerating}
Charlie Chen, Sebastian Borgeaud, Geoffrey Irving, Jean-Baptiste Lespiau, Laurent Sifre, and John Jumper. 2023{\natexlab{a}}.
\newblock Accelerating large language model decoding with speculative sampling.
\newblock \emph{arXiv preprint arXiv:2302.01318}.

\bibitem[{Chen et~al.(2023{\natexlab{b}})Chen, Zaharia, and Zou}]{chen2023frugalgpt}
Lingjiao Chen, Matei Zaharia, and James Zou. 2023{\natexlab{b}}.
\newblock Frugalgpt: How to use large language models while reducing cost and improving performance.
\newblock \emph{arXiv preprint arXiv:2305.05176}.

\bibitem[{Cobbe et~al.(2021)Cobbe, Kosaraju, Bavarian, Chen, Jun, Kaiser, Plappert, Tworek, Hilton, Nakano et~al.}]{cobbe2021training}
Karl Cobbe, Vineet Kosaraju, Mohammad Bavarian, Mark Chen, Heewoo Jun, Lukasz Kaiser, Matthias Plappert, Jerry Tworek, Jacob Hilton, Reiichiro Nakano, and 1 others. 2021.
\newblock Training verifiers to solve math word problems.
\newblock \emph{arXiv preprint arXiv:2110.14168}.

\bibitem[{Dathathri et~al.(2019)Dathathri, Madotto, Lan, Hung, Frank, Molino, Yosinski, and Liu}]{dathathri2019plug}
Sumanth Dathathri, Andrea Madotto, Janice Lan, Jane Hung, Eric Frank, Piero Molino, Jason Yosinski, and Rosanne Liu. 2019.
\newblock Plug and play language models: A simple approach to controlled text generation.
\newblock \emph{arXiv preprint arXiv:1912.02164}.

\bibitem[{Geifman and El-Yaniv(2017)}]{geifman2017selective}
Yonatan Geifman and Ran El-Yaniv. 2017.
\newblock Selective classification for deep neural networks.
\newblock In \emph{Advances in Neural Information Processing Systems}.

\bibitem[{Geva et~al.(2021)Geva, Khashabi, Segal, Khot, Roth, and Berant}]{geva2021did}
Mor Geva, Daniel Khashabi, Elad Segal, Tushar Khot, Dan Roth, and Jonathan Berant. 2021.
\newblock Did {A}ristotle use a laptop? {A} question answering benchmark with implicit reasoning strategies.
\newblock \emph{Transactions of the Association for Computational Linguistics}, 9:346--361.

\bibitem[{Grattafiori et~al.(2024)Grattafiori, Dubey, Jauhri, Pandey, Kadian, Al-Dahle, Letman, Mathur, Schelten, Vaughan et~al.}]{grattafiori2024llama}
Aaron Grattafiori, Abhimanyu Dubey, Abhinav Jauhri, Abhinav Pandey, Abhishek Kadian, Ahmad Al-Dahle, Aiesha Letman, Akhil Mathur, Alan Schelten, Alex Vaughan, and 1 others. 2024.
\newblock The {L}lama 3 herd of models.
\newblock \emph{arXiv preprint arXiv:2407.21783}.

\bibitem[{Hendrycks et~al.(2021)Hendrycks, Burns, Kadavath, Arora, Basart, Tang, Song, and Steinhardt}]{hendrycks2021measuring}
Dan Hendrycks, Collin Burns, Saurav Kadavath, Akul Arora, Steven Basart, Eric Tang, Dawn Song, and Jacob Steinhardt. 2021.
\newblock Measuring mathematical problem solving with the {MATH} dataset.
\newblock \emph{Advances in Neural Information Processing Systems (NeurIPS), Datasets and Benchmarks Track}.

\bibitem[{Ilharco et~al.(2022)Ilharco, Ribeiro, Wortsman, Gururangan, Schmidt, Hajishirzi, and Farhadi}]{ilharco2022editing}
Gabriel Ilharco, Marco~Tulio Ribeiro, Mitchell Wortsman, Suchin Gururangan, Ludwig Schmidt, Hannaneh Hajishirzi, and Ali Farhadi. 2022.
\newblock Editing models with task arithmetic.
\newblock \emph{arXiv preprint arXiv:2212.04089}.

\bibitem[{Kadavath et~al.(2022)Kadavath, Conerly, Askell, Henighan, Drain, Perez, Schiefer, Hatfield-Dodds, DasSarma, Tran-Johnson et~al.}]{kadavath2022language}
Saurav Kadavath, Tom Conerly, Amanda Askell, Tom Henighan, Dawn Drain, Ethan Perez, Nicholas Schiefer, Zac Hatfield-Dodds, Nova DasSarma, Eli Tran-Johnson, and 1 others. 2022.
\newblock Language models (mostly) know what they know.
\newblock \emph{arXiv preprint arXiv:2207.05221}.

\bibitem[{Kojima et~al.(2022)Kojima, Gu, Reid, Matsuo, and Iwasawa}]{kojima2022large}
Takeshi Kojima, Shixiang~Shane Gu, Machel Reid, Yutaka Matsuo, and Yusuke Iwasawa. 2022.
\newblock Large language models are zero-shot reasoners.
\newblock \emph{Advances in neural information processing systems}, 35:22199--22213.

\bibitem[{Kong et~al.(2022)Kong, Wang, Yu, and Zhang}]{kong2022accelerating}
Jun Kong, Jin Wang, Liang-Chih Yu, and Xuejie Zhang. 2022.
\newblock Accelerating inference for pretrained language models by unified multi-perspective early exiting.
\newblock In \emph{Proceedings of the 29th International Conference on Computational Linguistics}, pages 4677--4686.

\bibitem[{Krause et~al.(2021)Krause, Gotmare, McCann, Keskar, Joty, Socher, and Rajani}]{krause2021gedi}
Ben Krause, Akhilesh~Deepak Gotmare, Bryan McCann, Nitish~Shirish Keskar, Shafiq Joty, Richard Socher, and Nazneen~Fatema Rajani. 2021.
\newblock Gedi: Generative discriminator guided sequence generation.
\newblock In \emph{Findings of the Association for Computational Linguistics: EMNLP 2021}, pages 4929--4952.

\bibitem[{Leviathan et~al.(2023)Leviathan, Kalman, and Matias}]{leviathan2023fast}
Yaniv Leviathan, Matan Kalman, and Yossi Matias. 2023.
\newblock Fast inference from transformers via speculative decoding.
\newblock In \emph{International Conference on Machine Learning}, pages 19274--19286. PMLR.

\bibitem[{Li and Liang(2021)}]{li2021prefix}
Xiang~Lisa Li and Percy Liang. 2021.
\newblock Prefix-tuning: Optimizing continuous prompts for generation.
\newblock In \emph{Proceedings of the 59th Annual Meeting of the Association for Computational Linguistics and the 11th International Joint Conference on Natural Language Processing (Volume 1: Long Papers)}, pages 4582--4597.

\bibitem[{Li et~al.(2025)Li, Dong, Ma, Zhang, Jia, and Sui}]{li2025selfbudgeter}
Zheng Li, Qingxiu Dong, Jingyuan Ma, Di~Zhang, Kai Jia, and Zhifang Sui. 2025.
\newblock Selfbudgeter: Adaptive token allocation for efficient llm reasoning.
\newblock \emph{arXiv preprint arXiv:2505.11274}.

\bibitem[{Papineni et~al.(2002)Papineni, Roukos, Ward, and Zhu}]{papineni2002bleu}
Kishore Papineni, Salim Roukos, Todd Ward, and Wei-Jing Zhu. 2002.
\newblock {BLEU}: a method for automatic evaluation of machine translation.
\newblock In \emph{Proceedings of the 40th Annual Meeting of the Association for Computational Linguistics}, pages 311--318.

\bibitem[{Schuster et~al.(2022)Schuster, Fisch, Gupta, Dehghani, Bahri, Tran, Tay, and Metzler}]{schuster2022confident}
Tal Schuster, Adam Fisch, Jai Gupta, Mostafa Dehghani, Dara Bahri, Vinh~Q Tran, Yi~Tay, and Donald Metzler. 2022.
\newblock Confident adaptive language modeling.
\newblock In \emph{Advances in Neural Information Processing Systems}.

\bibitem[{Song et~al.(2025)Song, Wei, Luo, Fan, Liu, Wang, and Wang}]{song2025well}
Feifan Song, Shaohang Wei, Wen Luo, Yuxuan Fan, Tianyu Liu, Guoyin Wang, and Houfeng Wang. 2025.
\newblock Well begun is half done: Low-resource preference alignment by weak-to-strong decoding.
\newblock \emph{arXiv preprint arXiv:2506.07434}.

\bibitem[{Sun et~al.(2025)Sun, Zhao, Wei, Zhang, and Gong}]{sun2025well}
Yiliu Sun, Zicheng Zhao, Yang Wei, Yanfang Zhang, and Chen Gong. 2025.
\newblock Well begun, half done: Reinforcement learning with prefix optimization for {LLM} reasoning.
\newblock \emph{arXiv preprint arXiv:2512.15274}.

\bibitem[{Turner et~al.(2023)Turner, Thiergart, Leech, Udell, Vazquez, Mini, and MacDiarmid}]{turner2023steering}
Alexander~Matt Turner, Lisa Thiergart, Gavin Leech, David Udell, Juan~J Vazquez, Ulisse Mini, and Monte MacDiarmid. 2023.
\newblock Steering language models with activation engineering.
\newblock \emph{arXiv preprint arXiv:2308.10248}.

\bibitem[{Wang et~al.(2022)Wang, Wei, Schuurmans, Le, Chi, Narang, Chowdhery, and Zhou}]{wang2022self}
Xuezhi Wang, Jason Wei, Dale Schuurmans, Quoc Le, Ed~Chi, Sharan Narang, Aakanksha Chowdhery, and Denny Zhou. 2022.
\newblock Self-consistency improves chain of thought reasoning in language models.
\newblock \emph{arXiv preprint arXiv:2203.11171}.

\bibitem[{Wei et~al.(2022)Wei, Wang, Schuurmans, Bosma, Xia, Chi, Le, Zhou et~al.}]{wei2022chain}
Jason Wei, Xuezhi Wang, Dale Schuurmans, Maarten Bosma, Fei Xia, Ed~Chi, Quoc~V Le, Denny Zhou, and 1 others. 2022.
\newblock Chain-of-thought prompting elicits reasoning in large language models.
\newblock \emph{Advances in neural information processing systems}, 35:24824--24837.

\bibitem[{Yang et~al.(2024)Yang, Yang, Zhang, Hui, Zheng, Yu, Li, Liu, Huang, Wei, Lin, Yang, Tu, Zhang, Yang, Yang, Zhou, Lin, Dang, Lu, Bao, Yang, Yu, Li, Xue, Zhang, Zhu, Men, Lin, Li, Tang, Xia, Ren, Ren, Fan, Su, Zhang, Wan, Liu, Cui, Zhang, Qiu et~al.}]{qwen2025technical}
An~Yang, Baosong Yang, Beichen Zhang, Binyuan Hui, Bo~Zheng, Bowen Yu, Chengyuan Li, Dayiheng Liu, Fei Huang, Haoran Wei, Huan Lin, Jian Yang, Jianhong Tu, Jianwei Zhang, Jianxin Yang, Jiaxi Yang, Jingren Zhou, Junyang Lin, Kai Dang, and 24 others. 2024.
\newblock Qwen2.5 technical report.
\newblock \emph{arXiv preprint arXiv:2412.15115}.

\bibitem[{Yang et~al.(2025)Yang, Ma, Lin, and Wei}]{yang2025towards}
Wenkai Yang, Shuming Ma, Yankai Lin, and Furu Wei. 2025.
\newblock Towards thinking-optimal scaling of test-time compute for llm reasoning.
\newblock \emph{arXiv preprint arXiv:2502.18080}.

\bibitem[{Yao et~al.(2023)Yao, Yu, Zhao, Shafran, Griffiths, Cao, and Narasimhan}]{yao2023tree}
Shunyu Yao, Dian Yu, Jeffrey Zhao, Izhak Shafran, Tom Griffiths, Yuan Cao, and Karthik Narasimhan. 2023.
\newblock Tree of thoughts: Deliberate problem solving with large language models.
\newblock \emph{Advances in neural information processing systems}, 36:11809--11822.

\bibitem[{Zhao et~al.(2021)Zhao, Wallace, Feng, Klein, and Singh}]{zhao2021calibrate}
Zihao Zhao, Eric Wallace, Shi Feng, Dan Klein, and Sameer Singh. 2021.
\newblock Calibrate before use: Improving few-shot performance of language models.
\newblock In \emph{International conference on machine learning}, pages 12697--12706. PMLR.

\bibitem[{Zhou et~al.(2022)Zhou, Sch{\"a}rli, Hou, Wei, Scales, Wang, Schuurmans, Cui, Bousquet, Le et~al.}]{zhou2022least}
Denny Zhou, Nathanael Sch{\"a}rli, Le~Hou, Jason Wei, Nathan Scales, Xuezhi Wang, Dale Schuurmans, Claire Cui, Olivier Bousquet, Quoc Le, and 1 others. 2022.
\newblock Least-to-most prompting enables complex reasoning in large language models.
\newblock \emph{arXiv preprint arXiv:2205.10625}.

\end{thebibliography}


\appendix

\providecommand{\DO}{\textsc{draft-only}}
\providecommand{\TO}{\textsc{target-only}}
\providecommand{\OL}{\textsc{open-loop}}
\providecommand{\CL}{\textsc{closed-loop}}
\providecommand{\SD}{\textsc{spec-decode}}
\providecommand{\CAS}{\textsc{cascade}}
\providecommand{\PH}{\textsc{posthoc}}
\providecommand{\SC}{\textsc{self-consistency}}

\section{Appendix Roadmap}
\label{app:roadmap}

The appendix is organized around the main claims. \Cref{app:hparams,app:algorithms} give implementation details and decoding procedures. \Cref{app:features,app:reliability} describe the trajectory features and the reliability of the learned trigger as the observed prefix grows. \Cref{app:sweeps,app:nonmonotonic_interventions,app:attention_dynamics,app:q9_full} provide additional evidence on schedule sensitivity, intervention count, prefix dynamics, and cross-task generalization. \Cref{app:target_behavior} analyzes target-side intervention quality. \Cref{app:followability} reports an auxiliary draft-tuning experiment for target followability. \Cref{app:q15} gives the extended speculative-decoding comparison. \Cref{app:datasets,app:variance,app:limitations} summarize evaluation, statistical reporting, compute, and limitations.

\section{Implementation Details}
\label{app:hparams}

\Cref{tab:hparams} lists the default decoding, triggering, and predictor-training settings. Unless otherwise stated, draft decoding is greedy. Closed-loop SCD uses bounded target-generated text segments and a lightweight trigger trained only from draft-side trajectory statistics and final-answer correctness labels.

\begin{table}[t]
\centering
\caption{Default hyperparameters for sparse collaborative decoding.}
\label{tab:hparams}
\small
\setlength{\tabcolsep}{2.5pt}
\begin{tabular}{lll}
\toprule
Knob & Default & Main overrides \\
\midrule
\multicolumn{3}{l}{\emph{Decoding}} \\
Open-loop prefix \(K\) & \(32\) & open-loop sweep \\
Segment length \(J\) & \(16\) & \(J\in\{8,32,64,96\}\) \\
Max new tokens & \(1280\) & StrategyQA: \(256\) \\
Sampling & greedy & \shortstack[l]{\SC: \(T{=}0.7\),\\ top-\(p{=}0.95\), \(k{=}5\)} \\
\midrule
\multicolumn{3}{l}{\emph{Trigger and budget}} \\
Warmup \(W\) & \(16\) & StrategyQA: \(4\) \\
Cooldown \(C\) & \(8\) & StrategyQA: \(4\) \\
Call budget \(B\) & \(3\) & closed-loop sweeps \\
Threshold \(\theta\) & \(0.5\) & \(\theta\in[0.2,0.8]\) \\
Velocity window \(\Delta\) & \(3\) & fixed \\
\midrule
\multicolumn{3}{l}{\emph{Predictor training}} \\
Calibration size & \(400\) & StrategyQA: \(200\) \\
Feature dimension & \(85\) & fixed \\
L2 regularization & \(5{\times}10^{-3}\) & fixed \\
CV folds & \(5\) & fixed \\
Checkpoints & \(16\)--\(256\), full & fixed \\
\bottomrule
\end{tabular}
\end{table}

\section{Decoding Procedures and Baselines}
\label{app:algorithms}

All SCD variants use the same text-injection primitive: a target segment is appended to the visible context and the draft resumes generation. The variants differ only in call placement. \OL{} fixes a global insertion schedule, whereas \CL{} evaluates a learned failure predictor online.

\vspace{2pt}
\noindent\textbf{Closed-loop SCD.}
Given prompt \(x\), draft \(D\), target \(T\), predictor \(g_\psi\), threshold \(\theta\), segment length \(J\), warmup \(W\), cooldown \(C\), and call budget \(B\):
\begin{enumerate}
    \setlength{\itemsep}{1pt}
    \setlength{\topsep}{2pt}
    \item Initialize trajectory \(y\), call counter \(n_{\mathrm{calls}}=0\), and last-fire position \(t_{\mathrm{last}}=-\infty\).
    \item Generate the next draft token \(y_t\) and record draft-side signals \(s_t\), including entropy, NLL, margin, top-\(k\) mass, and their history.
    \item If \(t<W\), \(t-t_{\mathrm{last}}<C\), or \(n_{\mathrm{calls}}\ge B\), continue draft decoding.
    \item Otherwise compute trajectory features \(\phi_t=\phi(s_{1:t})\) and risk score \(r_t=g_\psi(\phi_t)\).
    \item If \(r_t\ge\theta\), ask the target to generate a bounded segment \(z_{1:J'}\sim T(\cdot\mid x,y)\), \(J'\le J\), append it to the visible context, and resume draft decoding.
    \item Stop when the draft emits EOS or the maximum generation length is reached.
\end{enumerate}

\vspace{2pt}
\noindent\textbf{Open-loop SCD.}
Open-loop SCD removes the learned trigger. The draft first generates a fixed prefix of length \(P\). The target then generates \(K\) tokens conditioned on the prompt and prefix. The generated target segment is appended to the context, and the draft completes the trajectory.

\vspace{2pt}
\noindent\textbf{Cascade and posthoc routing.}
Cascade and posthoc use the same trajectory-shape predictor as closed-loop SCD, but defer its use until after draft completion. Cascade sends predicted hard examples to the target as full-example queries. Posthoc rewriting asks the target to revise the completed draft. These baselines test whether the predictor is useful only as an example-level router or whether mid-trajectory intervention provides additional leverage.

\section{Trajectory Features and Predictor Training}
\label{app:features}

For each base series, we compute aggregate statistics over the observed prefix. \Cref{tab:full_feature_list} summarizes the resulting \(85\)-dimensional feature vector: \(12\) base series times \(7\) aggregates, plus trajectory length.

\begin{table}[t]
\centering
\caption{Trajectory-shape features used by the failure predictor. For each base series, we compute the same seven prefix-level aggregates; adding trajectory length gives \(12\times 7+1=85\) features.}
\label{tab:full_feature_list}
\small
\setlength{\tabcolsep}{3pt}
\begin{tabular}{@{}p{0.30\linewidth}p{0.62\linewidth}@{}}
\toprule
Component & Features \\
\midrule
Base series &
\(\{H,\mathrm{NLL},M,vH,v\mathrm{NLL},vM,\)
\(\ |vH|,|v\mathrm{NLL}|,|vM|,p_1,\log M,\mathrm{top5}\}\) \\
Aggregates &
\(\{\max,\min,\mathrm{mean},p10,\)
\(\ p90,\mathrm{std},\mathrm{slope}\}\) \\
Scalar &
trajectory length \\
\midrule
Total &
\(12\times 7+1=85\) \\
\bottomrule
\end{tabular}
\end{table}

\vspace{2pt}
\noindent\textbf{Predictor training.}
For each draft-dataset pair, we decode a calibration set with the draft model, record per-step uncertainty statistics and final-answer correctness, sample truncated prefixes from each trajectory, standardize features, and train an L2-regularized logistic predictor with five-fold cross-validation. The predictor uses only draft-side traces and final correctness labels. It does not use target outputs or intervention outcomes, which separates trigger learning from target behavior.

\vspace{2pt}
\noindent\textbf{Feature interpretation.}
For Qwen-1.5B on GSM8K, the largest standardized weights are dominated by log-margin slope, margin-velocity variation, top-5 mass statistics, high entropy percentiles, and high NLL percentiles. This supports the main result: failures are marked by declining separation, increased volatility, and weaker probability concentration rather than isolated uncertainty spikes.

\section{Predictor Reliability Over Trajectory Length}
\label{app:reliability}

\begin{figure}[t]
  \centering
  \includegraphics[width=\linewidth]{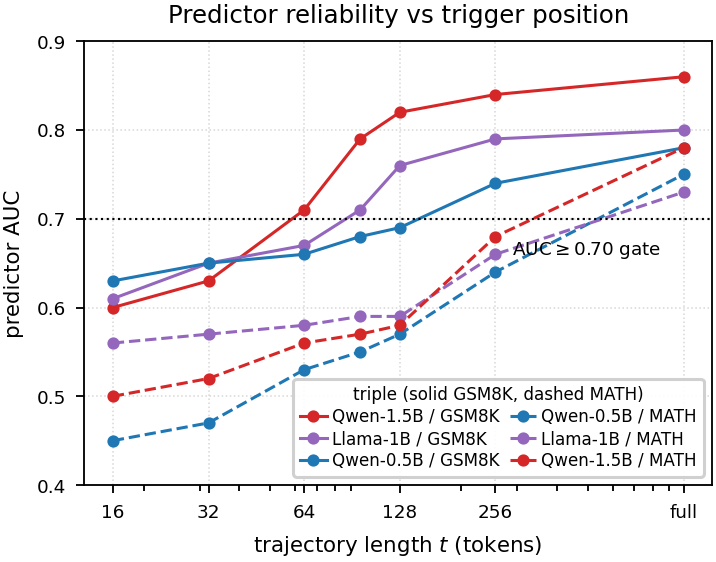}
  \caption{\textbf{Predictor reliability improves with trajectory length.}
  GSM8K predictors cross the operational \(\textsc{auc}=0.70\) gate earlier, while MATH-500 predictors remain weaker at comparable trigger positions. This explains why the same closed-loop mechanism is more reliable on GSM8K and more target-dependent on MATH-500.}
  \label{fig:app_auc_t}
\end{figure}

\Cref{fig:app_auc_t,tab:reliability_full} summarize how failure-prediction reliability changes with available trajectory length. The main deployment rule requires a trigger that is reliable early enough to act, not merely accurate on completed trajectories. MATH-500 reaches moderate full-trajectory AUC, but remains weak at early trigger points; by the time sufficient evidence accumulates, the draft may already have committed to an incorrect symbolic path.

\begin{table}[t]
\centering
\caption{Failure-prediction AUC at different prefix lengths. Bold marks the operational online-triggering point used in the main analysis.}
\label{tab:reliability_full}
\small
\setlength{\tabcolsep}{1.5pt}
\begin{tabular}{llrrrrrrr}
\toprule
Draft & Data & @16 & @32 & @64 & @96 & @128 & @256 & Full \\
\midrule
Qwen 1.5B & GSM8K & 0.60 & 0.63 & 0.71 & 0.79 & \textbf{0.82} & 0.84 & 0.86 \\
Llama 1B & GSM8K & 0.61 & 0.65 & 0.67 & 0.71 & \textbf{0.76} & 0.79 & 0.80 \\
Qwen 0.5B & GSM8K & 0.63 & 0.65 & 0.66 & 0.68 & \textbf{0.69} & 0.74 & 0.78 \\
\midrule
Qwen 0.5B & MATH & 0.45 & 0.47 & 0.53 & 0.55 & 0.57 & 0.64 & 0.75 \\
Llama 1B & MATH & 0.56 & 0.57 & 0.58 & 0.59 & 0.59 & 0.66 & 0.73 \\
Qwen 1.5B & MATH & 0.50 & 0.52 & 0.56 & 0.57 & 0.58 & 0.68 & 0.78 \\
\bottomrule
\end{tabular}
\end{table}

\section{Open-Loop and Closed-Loop Sweeps}
\label{app:sweeps}

\begin{figure}[t]
  \centering
  \includegraphics[width=\linewidth]{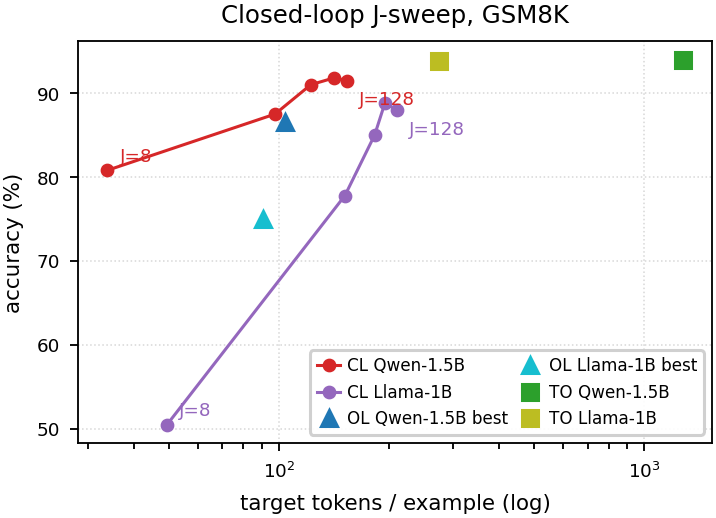}
  \caption{\textbf{Closed-loop accuracy-cost frontier.}
  Increasing segment length \(J\) improves closed-loop SCD until target use saturates. The learned trigger improves over the best open-loop points, especially for the weaker Llama-1B draft where more errors remain recoverable.}
  \label{fig:app_cl_jsweep}
\end{figure}

\begin{figure}[t]
  \centering
  \includegraphics[width=\linewidth]{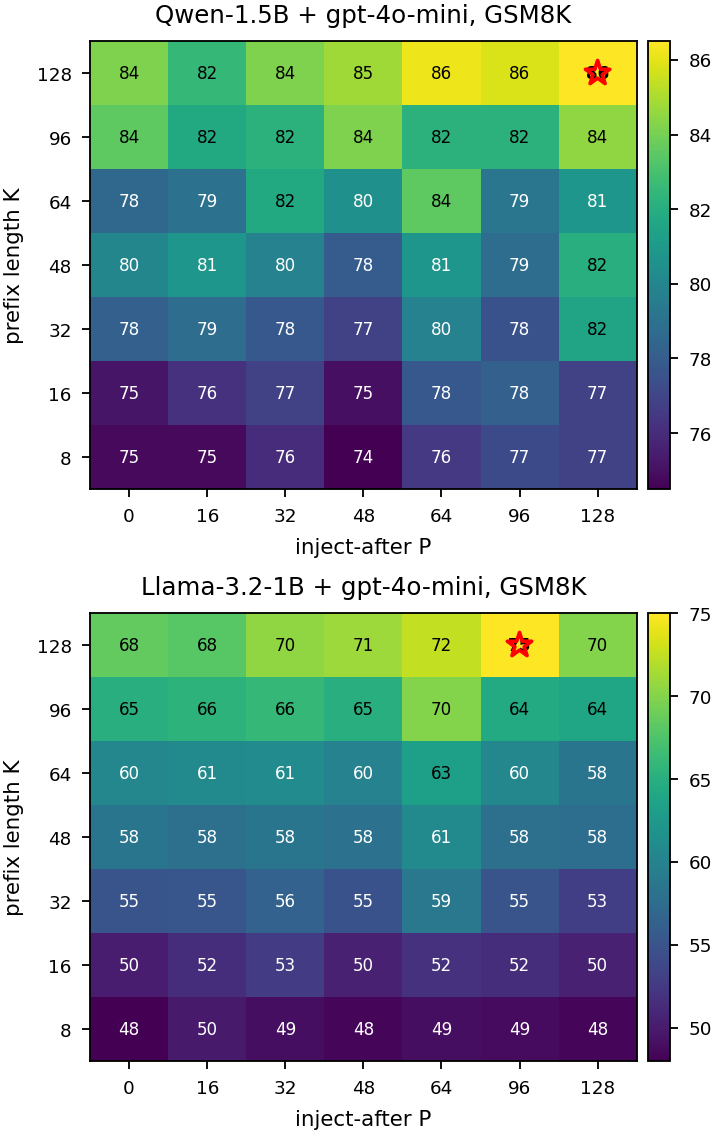}
  \caption{\textbf{Open-loop SCD depends on insertion position and segment length.}
  Fixed schedules can improve accuracy, but the best insertion point depends on the draft and schedule. Learned triggering avoids committing to one global schedule across examples and instead adapts to trajectory-specific drift.}
  \label{fig:app_openloop_heatmap}
\end{figure}

\Cref{fig:app_cl_jsweep,tab:cl_j_full,fig:app_openloop_heatmap,tab:posdep_full} provide the full sweep results behind the compact main-text tables. They show why open-loop SCD is a strong baseline and why learned triggering improves over fixed schedules. Gains are not caused by simply adding more target tokens: very short target segments are too weak to repair the trajectory, while longer segments saturate and can increase overwrite risk.

\begin{table}[t]
\centering
\caption{Full closed-loop \(J\)-sweep on GSM8K. Rsc = rescue rate, Brk = break rate, Tok = target tokens per example, Cls = target calls per example.}
\label{tab:cl_j_full}
\small
\setlength{\tabcolsep}{3pt}
\begin{tabular}{lrrrrrr}
\toprule
Draft & \(J\) & Acc & Rsc & Brk & Tok & Cls \\
\midrule
Qwen 1.5B & 8 & 80.8 & 10.8 & 4.5 & 33.9 & 1.21 \\
Qwen 1.5B & 32 & 87.5 & 15.2 & 2.2 & 97.6 & 1.78 \\
Qwen 1.5B & 64 & 91.0 & 17.5 & 1.0 & 122.3 & 2.06 \\
Qwen 1.5B & 96 & \textbf{91.8} & 18.2 & 1.0 & 141.3 & 2.18 \\
Qwen 1.5B & 128 & 91.5 & 18.0 & 1.0 & 153.8 & 2.22 \\
\midrule
Llama 1B & 8 & 50.5 & 11.8 & 9.0 & 49.4 & 2.55 \\
Llama 1B & 32 & 77.8 & 33.2 & 3.2 & 151.7 & 2.49 \\
Llama 1B & 64 & 85.0 & 40.0 & 2.8 & 182.9 & 2.32 \\
Llama 1B & 96 & \textbf{88.8} & 42.5 & 1.5 & 195.1 & 2.27 \\
Llama 1B & 128 & 88.0 & 42.0 & 1.8 & 210.6 & 2.21 \\
\bottomrule
\end{tabular}
\end{table}

\begin{table}[t]
\centering
\caption{Position-dependent threshold sweep at \(J=96\) on GSM8K. The best setting is mildly position-dependent for Qwen-1.5B and more strongly threshold-sensitive for Llama-1B.}
\label{tab:posdep_full}
\small
\setlength{\tabcolsep}{3.5pt}
\begin{tabular}{lccccc}
\toprule
Draft / \(\theta_e\) & \multicolumn{5}{c}{\(\theta_l\)} \\
\cmidrule(lr){2-6}
 & 0.2 & 0.3 & 0.5 & 0.7 & 0.8 \\
\midrule
\multicolumn{6}{l}{\emph{Qwen-1.5B}} \\
0.3 & 91.2 & 91.5 & 91.0 & 90.0 & 88.5 \\
0.5 & 91.8 & \textbf{92.2} & 90.8 & 88.8 & 87.2 \\
0.7 & 90.5 & 90.8 & 89.5 & 88.0 & 86.0 \\
\midrule
\multicolumn{6}{l}{\emph{Llama-1B}} \\
0.3 & 89.2 & 89.8 & \textbf{90.0} & 84.5 & 80.0 \\
0.5 & 87.0 & 87.5 & 86.0 & 81.0 & 78.0 \\
0.7 & 80.5 & 80.0 & 77.5 & 77.2 & 73.0 \\
\bottomrule
\end{tabular}
\end{table}

\section{Non-Monotonic Accuracy Under Increasing Target Interventions}
\label{app:nonmonotonic_interventions}

\begin{figure*}[t]
  \centering
  \includegraphics[width=\textwidth]{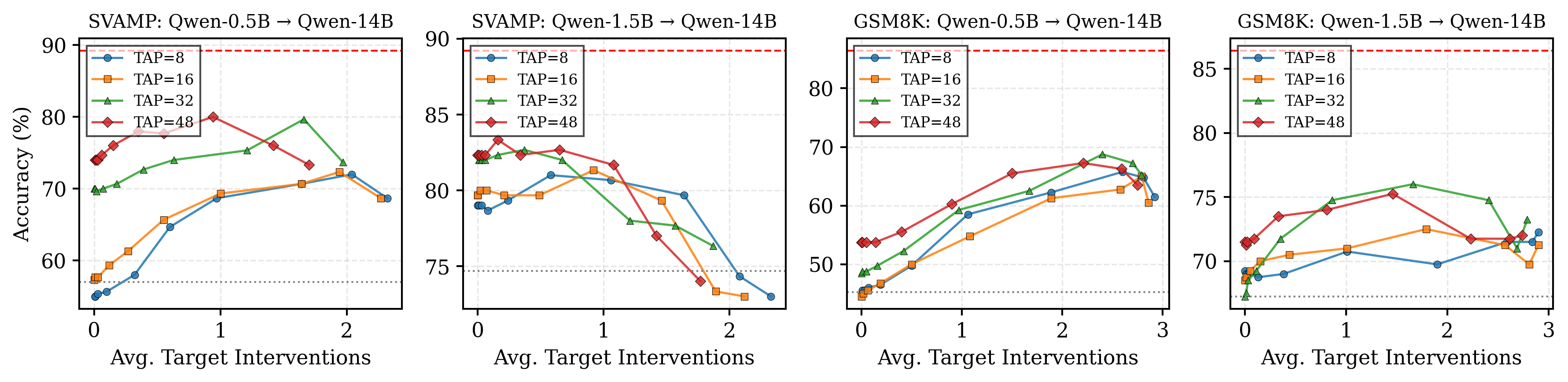}
  \caption{\textbf{Accuracy versus average target interventions.}
  Results are shown for SVAMP and GSM8K with Qwen-0.5B and Qwen-1.5B drafts and a fixed Qwen-14B target. Curves correspond to different target-assisted prefix (TAP) lengths, where \texttt{TAP=$K$} injects the first \(K\) target tokens before draft decoding resumes. Accuracy improves with a small number of interventions, then saturates or declines when intervention becomes too aggressive.}
  \label{fig:nonmonotonic_interventions}
\end{figure*}

\Cref{fig:nonmonotonic_interventions} shows that accuracy is not monotonic in the average number of target interventions per example. A small number of well-timed interventions can redirect unstable trajectories, while excessive intervention can overwrite useful draft state or fire after the controllable window has passed.

TAP is an open-loop prefixing mechanism: the first \(K\) target-generated tokens are inserted before draft decoding continues, with \(K\in\{8,16,32,48\}\). The x-axis in \Cref{fig:nonmonotonic_interventions} is the average number of target interventions per example, not a fixed per-example budget. It emerges from the trigger threshold, the evolving draft trajectory, and problem difficulty. Lower thresholds make the controller more aggressive.

Across settings, moving from zero or very few interventions to a small number improves accuracy. After this point, additional calls have diminishing or negative returns: some fire after the draft has committed, and others inject target text into trajectories that were already correct or recoverable. TAP length and intervention frequency therefore control different parts of the same process. TAP shapes trajectory initialization, while closed-loop SCD controls when to reallocate target computation during decoding.

\section{Attention Dynamics under Target Prefixing}
\label{app:attention_dynamics}

This appendix provides mechanistic evidence for how target-generated prefixes reshape draft attention dynamics. The goal is not to show copying, but to understand when prefix intervention changes reasoning commitment, attention concentration, and trajectory anchoring.

\vspace{2pt}
\noindent\textbf{Setup.}
All experiments use GSM8K with greedy decoding. The target is Qwen2.5-14B-Instruct, and the drafts are Qwen2.5-1.5B-Instruct and Qwen2.5-0.5B-Instruct. Attention statistics are computed on 100 GSM8K examples and averaged over 8 transformer layers. Generation length is fixed to 512 tokens. We analyze \(K=48\) and \(K=96\), representing moderate and stronger anchoring.

\vspace{2pt}
\noindent\textbf{Attention entropy.}
We measure per-token attention entropy
\[
H = -\sum_i p_i \log p_i,
\]
where \(p_i\) denotes normalized attention weights. Lower entropy indicates more concentrated attention.

\begin{table}[t]
\centering
\caption{Attention entropy over decoding phases. Lower entropy indicates stronger attention concentration.}
\label{tab:attention_entropy}
\small
\setlength{\tabcolsep}{4pt}
\begin{tabular}{lccc}
\toprule
Condition & Early & Mid & Late \\
 & 0--20 & 20--50 & 50--100 \\
\midrule
Target & 2.446 & 2.568 & 2.650 \\
Draft & 2.791 & 3.028 & 3.178 \\
Draft + K48 & 3.079 & 3.176 & 3.192 \\
Draft + K96 & 3.206 & 3.172 & 3.075 \\
\bottomrule
\end{tabular}
\end{table}

\Cref{tab:attention_entropy} shows that the draft alone has steadily increasing entropy, consistent with diffusion of attention over decoding. Prefixing with \(K=48\) stabilizes this trend but does not reverse it. In contrast, \(K=96\) produces a late-phase entropy decrease, suggesting more convergent attention dynamics rather than only a local prefix bias.

\vspace{2pt}
\noindent\textbf{Entropy slope.}
We summarize each trajectory by fitting a linear slope to attention entropy over decoding steps.

\begin{table}[t]
\centering
\caption{Linear attention entropy slope per token. Negative slope indicates increasing attention concentration over time.}
\label{tab:entropy_slopes}
\small
\setlength{\tabcolsep}{6pt}
\begin{tabular}{lc}
\toprule
Condition & Entropy slope \\
\midrule
Target & \(+0.00254\) \\
Draft & \(+0.00484\) \\
Draft + K48 & \(+0.00141\) \\
Draft + K96 & \(\mathbf{-0.00164}\) \\
\bottomrule
\end{tabular}
\end{table}

Only \(K=96\) yields a negative entropy slope in \Cref{tab:entropy_slopes}, indicating increasing attention concentration over time. This supports the interpretation that longer target prefixes can transfer useful trajectory structure, not merely local token content.

\vspace{2pt}
\noindent\textbf{Prefix attention dominance.}
To test whether the effect is persistent copying from the target prefix, we measure the fraction of attention mass assigned to prefix tokens.

\begin{table}[t]
\centering
\caption{Prefix attention dominance and decay over decoding.}
\label{tab:prefix_ratio}
\small
\setlength{\tabcolsep}{6pt}
\begin{tabular}{lccc}
\toprule
Condition & Early ratio & Late ratio & Decay \\
\midrule
Draft + K48 & 0.855 & 0.635 & \(-25.8\%\) \\
Draft + K96 & 0.863 & 0.651 & \(-24.6\%\) \\
\bottomrule
\end{tabular}
\end{table}

\Cref{tab:prefix_ratio} shows that both prefix lengths decay to similar late-stage prefix attention despite the longer \(K=96\) prefix. The effect is therefore not explained by persistent copying of target tokens. Target prefixing appears to shape the early trajectory state, after which the draft continues with altered attention dynamics.

\section{Full Cross-Task Generalization Grid}
\label{app:q9_full}

\Cref{tab:q9_full} reports the full cross-task grid summarized in the main paper. It separates broad GSM8K gains from the more conditional MATH-500 behavior.

\begin{table*}[t]
\centering
\caption{Full generalization grid. \(\Delta_{\mathrm{CL-OL}}\) is best closed-loop minus best open-loop accuracy in percentage points.}
\label{tab:q9_full}
\small
\setlength{\tabcolsep}{5pt}
\begin{tabular}{lllrrrr}
\toprule
Data & Draft & Target & \DO & best-\OL & best-\CL & \(\Delta_{\mathrm{CL-OL}}\) \\
\midrule
GSM8K & Qwen 0.5B & 4o-mini & 48.0 & 69.5 & \textbf{89.5} & \(+20.0\) \\
GSM8K & Qwen 0.5B & 4o & 48.0 & 72.5 & \textbf{91.0} & \(+18.5\) \\
GSM8K & Llama 1B & 4o & 45.0 & 68.5 & \textbf{86.5} & \(+18.0\) \\
GSM8K & Qwen 1.5B & 4o-mini & 74.5 & 86.5 & \textbf{91.8} & \(+5.3\) \\
GSM8K & Llama 1B & 4o-mini & 47.8 & 75.0 & \textbf{88.8} & \(+13.8\) \\
\midrule
MATH & Qwen 0.5B & 4o-mini & 24.5 & \textbf{37.5} & 32.0 & \(-5.5\) \\
MATH & Llama 1B & 4o-mini & 14.0 & 19.5 & \textbf{30.5} & \(+11.0\) \\
MATH & Llama 1B & 4o & 14.0 & 21.5 & \textbf{35.0} & \(+13.5\) \\
MATH & Qwen 1.5B & 4o-mini & 42.5 & \textbf{46.5} & 42.0 & \(-4.5\) \\
MATH & Qwen 0.5B & 4o & 24.5 & 34.5 & \textbf{41.5} & \(+7.0\) \\
\bottomrule
\end{tabular}
\end{table*}

The full grid reinforces two conclusions. First, the largest gains occur for weaker drafts, where more errors remain recoverable. Second, MATH-500 failures are selective rather than universal: the same draft can lose with a weaker or less bounded target but recover when paired with a target that can repair the fired context.

\section{Target Behavior and Intervention Reliability}
\label{app:target_behavior}

\begin{figure}[t]
  \centering
  \includegraphics[width=0.82\linewidth]{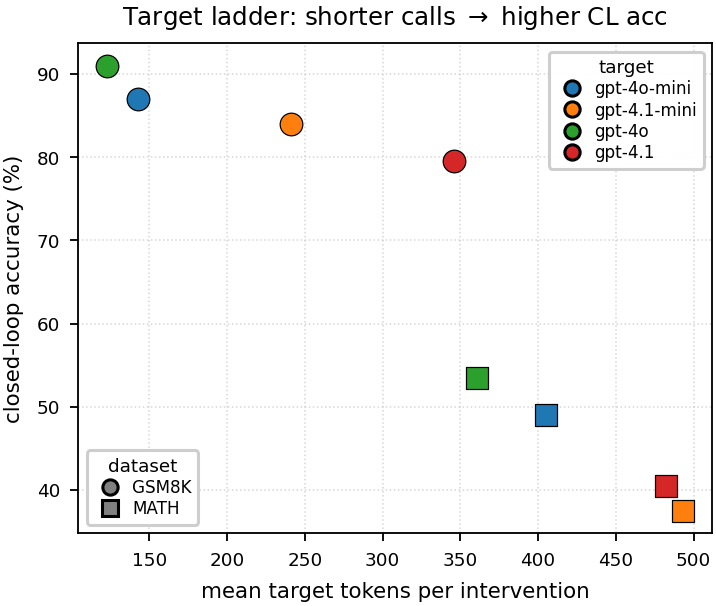}
  \caption{\textbf{Target intervention behavior matters beyond target-only accuracy.}
  Shorter target continuations are associated with better closed-loop accuracy. Stronger target-only models can degrade closed-loop performance when their continuations are long or overwrite the draft context.}
  \label{fig:app_target_ladder}
\end{figure}

The main paper shows that trigger reliability is only one part of successful sparse collaboration. Here we analyze the second requirement: whether the target continuation is bounded, confident, and compatible with the current draft context. \Cref{fig:app_target_ladder,tab:q13_full,tab:target_confidence2} show that target-only accuracy is not sufficient for good intervention behavior.

\begin{table*}[t]
\centering
\caption{Target ladder with rescue, degradation, and captured fraction. Cap is the fraction of the draft-to-target gap captured by closed-loop SCD.}
\label{tab:q13_full}
\small
\setlength{\tabcolsep}{3pt}
\begin{tabular}{llrrrrrr}
\toprule
Data & Target & \DO & \TO & \CL & Rsc & Deg & Cap \\
\midrule
GSM8K & 4o-mini & 73.5 & 94.5 & 87.0 & 15.0 & 1.5 & 0.64 \\
GSM8K & 4.1-mini & 73.5 & 97.5 & 84.0 & 15.0 & 4.5 & 0.44 \\
GSM8K & 4o & 73.5 & 95.0 & \textbf{91.0} & \textbf{19.5} & 2.0 & \textbf{0.81} \\
GSM8K & 4.1 & 73.5 & 96.5 & 79.5 & 13.0 & 7.0 & 0.26 \\
\midrule
MATH & 4o-mini & 42.5 & 69.5 & 49.0 & 11.5 & 5.0 & 0.24 \\
MATH & 4.1-mini & 42.5 & 72.0 & 37.5 & 7.0 & 12.0 & \(-0.17\) \\
MATH & 4o & 42.5 & 71.0 & \textbf{53.5} & \textbf{15.5} & 4.5 & \textbf{0.39} \\
MATH & 4.1 & 42.5 & 71.5 & 40.5 & 8.5 & 10.5 & \(-0.07\) \\
\bottomrule
\end{tabular}
\end{table*}

\begin{table*}[t]
\centering
\caption{Target-side confidence as an intervention quality signal. Metrics are computed over the injected target segment. Recovery cases correct an unstable or incorrect draft trajectory; degradation cases worsen an otherwise correct or recoverable trajectory.}
\label{tab:target_confidence2}
\small
\setlength{\tabcolsep}{5pt}
\begin{tabular}{lccc}
\toprule
Metric & Recovery & Degradation & Direction \\
\midrule
\multicolumn{4}{l}{\emph{Per-token averages during injection}} \\
Mean NLL \(\downarrow\) & 0.162 & 0.200 & higher is worse \\
Mean entropy \(\downarrow\) & 0.336 & 0.413 & higher is worse \\
Mean max prob \(\uparrow\) & 0.877 & 0.854 & lower is worse \\
Mean margin \(\uparrow\) & 0.795 & 0.765 & lower is worse \\
\midrule
\multicolumn{4}{l}{\emph{Uncertainty flags: fraction of interventions with at least one flagged token}} \\
High entropy \(>3.0\) & 0.2\% & 3.3\% & more frequent in degradation \\
High NLL \(>2.0\) & 0.0\% & 1.6\% & more frequent in degradation \\
\bottomrule
\end{tabular}
\end{table*}

The target ladder explains why target-only accuracy alone is not a reliable deployment criterion. Models with higher standalone accuracy can produce longer or more intrusive continuations, increasing degradation. \Cref{tab:target_confidence2} further shows that degradation cases have higher target uncertainty at injection time. This suggests that the target can be unreliable exactly when it is asked to repair the draft, and that target-side confidence can be used as an additional guardrail.

\vspace{2pt}
\noindent\textbf{Linguistic signatures and failure cases.}
Harmful interventions often contain hedging, meta-reasoning, or speculative phrasing rather than direct computation. Successful interventions are more often assertive and computation-focused, with explicit numeric operations and direct deductive transitions. Manual inspection of degrading trajectories shows three recurring modes: target continuations that overwrite a partially correct draft plan, late triggers after the draft has committed to a wrong setup, and answer-format mismatches in symbolic tasks. These failures motivate shorter target continuations, position-dependent thresholds, and task-specific answer normalization.

\section{Draft Tuning for Target Followability}
\label{app:followability}

This section is not used in any main comparison. It reports an auxiliary experiment that asks whether a draft can be lightly adapted to better continue externally injected target prefixes. This is orthogonal to the trigger policy: the trigger decides \emph{when} to inject target text, while followability tuning improves whether the draft can \emph{retain} the injected trajectory after control returns to the draft.

\vspace{2pt}
\noindent\textbf{Motivation and objective.}
Open-loop and closed-loop SCD rely on a target-to-draft handoff. Failures can arise even when the target segment is useful, if the draft later drifts away, overwrites the injected reasoning state, or fails to continue it coherently. Followability tuning is a lightweight offline adaptation that does not use task labels, final-answer correctness, rewards, or preference supervision. Given a prompt \(P\) and target-generated prefix \(T_{1:K}\), the draft is trained to treat the injected prefix as a valid trajectory state and continue coherently from it.

Let \(D_\theta\) denote the draft and \(T\) the target. For each prompt, we sample a target prefix \(T_{1:K}\) and train a small adapter on the draft using
\[
\mathcal{L}
=
\lambda_{\mathrm{inj}}\mathcal{L}_{\mathrm{inj}}
+
\lambda_{\mathrm{cont}}\mathcal{L}_{\mathrm{cont}} .
\]
Here, \(\mathcal{L}_{\mathrm{inj}}\) aligns the draft next-token distribution at the injection boundary, while \(\mathcal{L}_{\mathrm{cont}}\) aligns continuation behavior over subsequent tokens. The continuation loss is exponentially decayed with distance from the injection point.

\vspace{2pt}
\noindent\textbf{Adapter parameterization.}
We use low-rank adapters in the query and value projections of selected attention layers. The adapter parameters are less than \(0.02\%\) of the draft parameters. They are trained offline and frozen at inference time, adding no target calls and no target-side latency during decoding.

\begin{figure*}[t]
  \centering
  \includegraphics[width=0.4\linewidth, trim=0 0 420 0, clip]{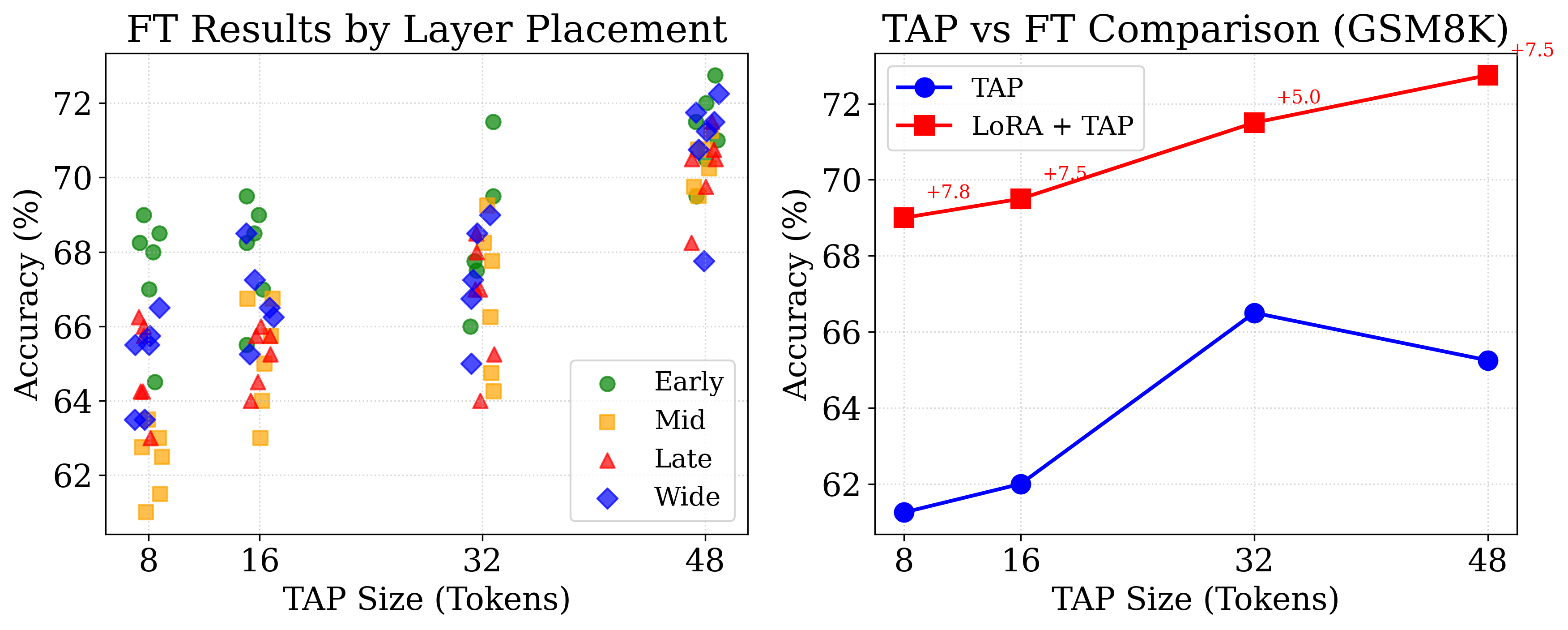}
  \includegraphics[width=0.4\linewidth]{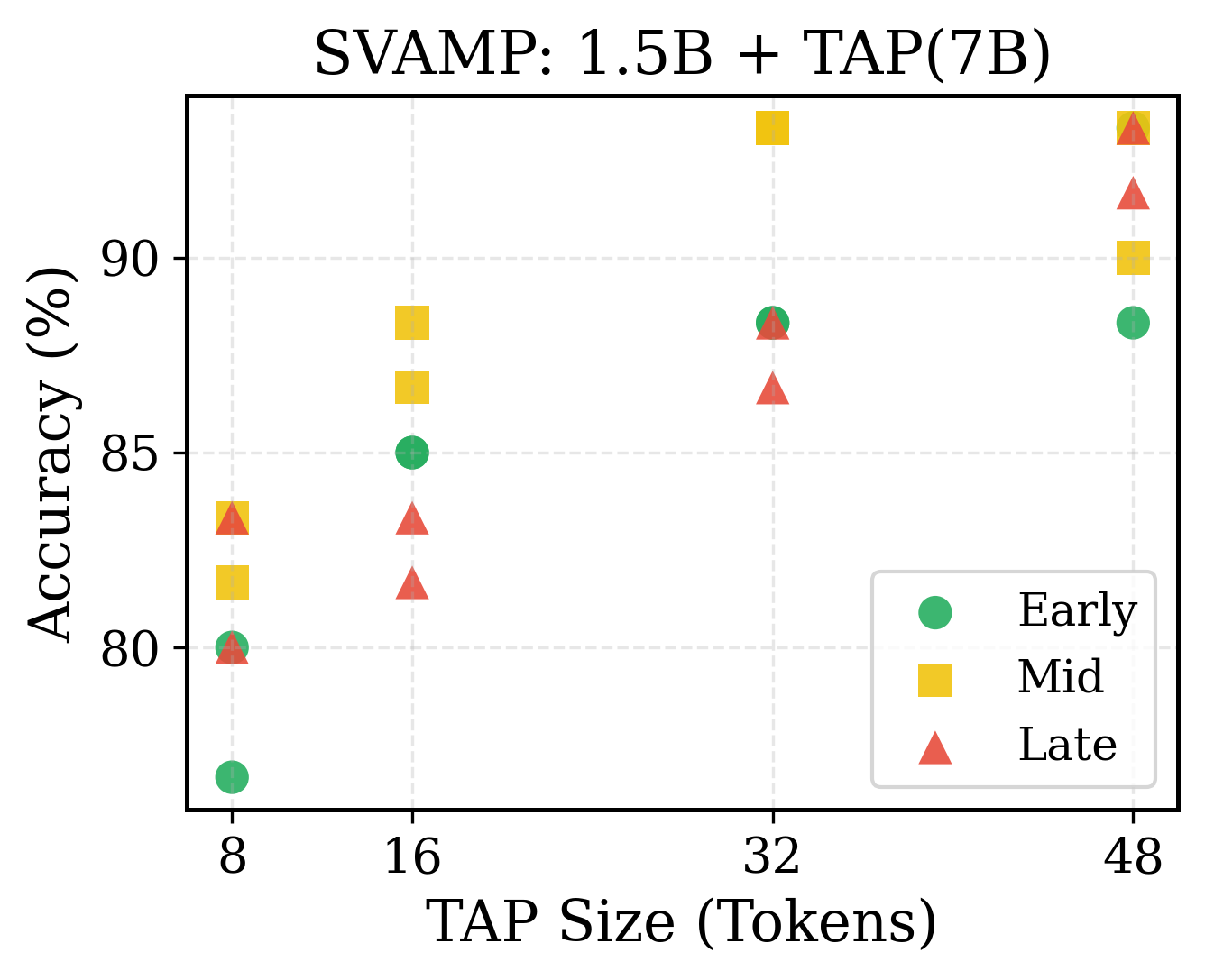}
  \caption{\textbf{Draft tuning for target followability.}
  Left: layer-wise sensitivity on GSM8K, where gains concentrate in early-to-mid layers, suggesting stabilization near the target-to-draft handoff. Right: layer-wise sensitivity on SVAMP, where gains peak in middle layers, suggesting improved longer-horizon retention of injected trajectories.}
  \label{fig:followability_layers}
\end{figure*}

\vspace{2pt}
\noindent\textbf{Results.}
\Cref{fig:followability_layers,tab:followability_main} summarize the auxiliary followability results. On GSM8K, gains concentrate in early-to-mid layers; on SVAMP, gains peak in middle layers, consistent with longer-horizon retention.

\begin{table}[t]
\centering
\caption{Followability sweep on GSM8K. Draft: Qwen2.5-1.5B-Instruct. Target: Qwen2.5-14B-Instruct. Training uses \(K_{\mathrm{target}}=80\); evaluation injects \(K_{\mathrm{draft}}=32\).}
\label{tab:followability_main}
\small
\setlength{\tabcolsep}{4pt}
\begin{tabular}{lcc}
\toprule
Config & Acc & Lat. (ms) \\
\midrule
No adapter & 0.6200 & 9497.3 \\
\midrule
early r8 L1+ent & \textbf{0.6650} & 10388.1 \\
early r8 L1 & 0.6575 & 10022.3 \\
early r4 L1+ent & 0.6425 & 9637.5 \\
early r4 L1 & 0.6400 & 9467.8 \\
\midrule
mid r8 L1+ent & 0.6450 & 9992.0 \\
mid r8 L1 & 0.6375 & 10082.2 \\
mid r4 L1+ent & 0.6325 & 9835.2 \\
mid r4 L1 & 0.6275 & 9642.3 \\
\midrule
late r8 L1+ent & 0.6375 & 9590.1 \\
late r8 L1 & 0.6300 & 9958.8 \\
late r4 L1+ent & 0.6275 & 9804.4 \\
late r4 L1 & 0.6250 & 10199.0 \\
\bottomrule
\end{tabular}
\end{table}

The best configuration improves GSM8K accuracy from 0.6200 to 0.6650. Early-layer adaptation is most effective, \(r=8\) generally outperforms \(r=4\), and entropy regularization is mildly beneficial for early and middle placements. These results suggest that the target-to-draft handoff is governed by early computations that incorporate the injected prefix into the draft state.

\section{Speculative Decoding Comparison}
\label{app:q15}

\begin{figure}[t]
  \centering
  \includegraphics[width=0.8\linewidth]{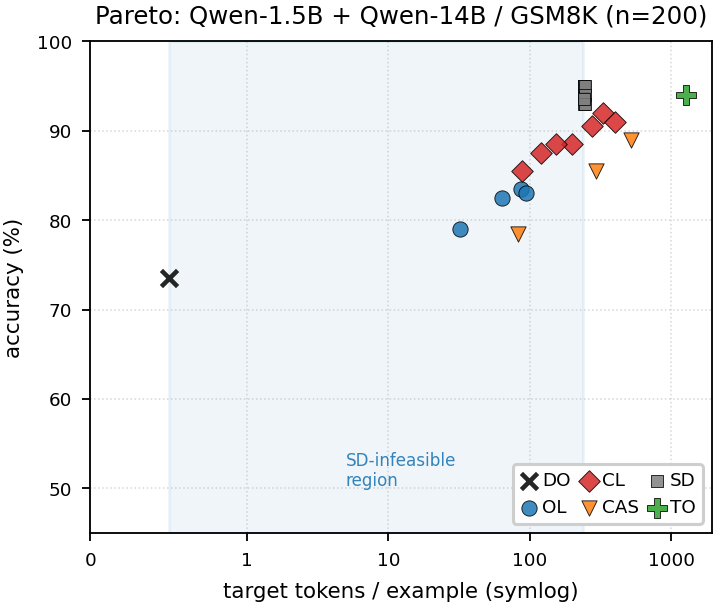}
  \caption{\textbf{Speculative decoding and SCD occupy different regimes.}
  In SD-feasible open-weight settings, speculative decoding provides the dense parity-preserving point. SCD provides sparse text-correction operating points and applies to black-box/API targets where target logits are unavailable.}
  \label{fig:app_pareto_q15}
\end{figure}

\Cref{fig:app_pareto_q15,tab:q15_pareto_full} summarize the SD-feasible open-weight comparison. These results place SCD relative to the closest dense draft-target baseline, but do not imply that SCD should replace SD when SD assumptions hold.

The table reports accuracy and mean target tokens per example for Qwen-family settings where SD is well-defined. MATH rows should be interpreted with the same exact-match extractor used throughout the paper; because symbolic formatting is more sensitive than GSM8K, the main conclusion is about deployment regime and target-token allocation rather than claiming an SD-over-target accuracy result.

\begin{table}[t]
\centering
\caption{Extended SD-feasible comparison on Qwen-family pairs. Tok is mean target tokens per example.}
\label{tab:q15_pareto_full}
\small
\setlength{\tabcolsep}{5pt}
\begin{tabular}{lrr}
\toprule
Method & Acc & Tok \\
\midrule
\multicolumn{3}{l}{\emph{Qwen-1.5B / GSM8K}} \\
\DO & 73.5 & 0 \\
\OL & 83.5 & 87 \\
\CL{} low & 87.5 & 120 \\
\CL{} best & 92.0 & 330 \\
\SD & 93.0--95.0 & 244 \\
\TO & 94.0 & 1280 \\
\midrule
\multicolumn{3}{l}{\emph{Qwen-0.5B / GSM8K}} \\
\DO & 48.0 & 0 \\
\OL & 69.5 & 109 \\
\CL & 88.5 & 409 \\
\CL{} best & 91.0 & 501 \\
\SD & 94.0--95.0 & 244 \\
\TO & 93.5 & 512 \\
\midrule
\multicolumn{3}{l}{\emph{Qwen-1.5B / MATH}} \\
\DO & 50.0 & 0 \\
\OL & 57.5 & 95 \\
\CL & 55.5 & 198 \\
\SD & 70.5--74.0 & 556 \\
\TO & 48.0 & 507 \\
\midrule
\multicolumn{3}{l}{\emph{Qwen-0.5B / MATH}} \\
\DO & 29.5 & 0 \\
\OL & 38.0 & 125 \\
\CL & 27.0 & 336 \\
\SD & 70.5--71.5 & 558 \\
\TO & 48.0 & 507 \\
\bottomrule
\end{tabular}
\end{table}

Where SD is feasible, it approaches target-level accuracy through dense verification. SCD should not be interpreted as an SD replacement in that regime. Instead, it provides sparse text-correction operating points and applies in black-box target settings where target logits are unavailable.

\section{Datasets, Prompting, and Evaluation}
\label{app:datasets}

\vspace{2pt}
\noindent\textbf{GSM8K.}
We use the standard test split and exact-match numeric answer extraction. Prompts ask the model to reason step by step and provide a final numeric answer.

\vspace{2pt}
\noindent\textbf{MATH-500.}
We evaluate exact-match answers extracted from boxed final responses. Because symbolic answers can have equivalent forms, MATH-500 results are more sensitive to answer extraction than GSM8K.

\vspace{2pt}
\noindent\textbf{StrategyQA.}
We evaluate 300 questions using a multi-hop yes/no prompting format. Answer extraction uses the final yes/no response.

\section{Statistical Reporting and Compute}
\label{app:variance}

All main cells use paired example lists across methods. Most cells use \(n=200\), with StrategyQA using \(n=300\). At \(n=200\), the binomial standard error is at most \(3.5\) percentage points at \(50\%\) accuracy and \(1.6\) points near \(90\%\). Larger closed-loop versus open-loop gains are unlikely to be explained by binomial noise alone; smaller differences should be interpreted cautiously. We report target tokens and calls as the primary cost measures, with wall-clock time included where available.

\end{document}